\RequirePackage{fix-cm}
\documentclass[twocolumn]{svjour3}          %

\smartqed  %
\usepackage{graphicx}
\usepackage{hyperref}
\hypersetup{
    colorlinks=true,
    breaklinks=true,
    linkcolor=red,
    filecolor=magenta,      
    urlcolor=blue,
    citecolor=blue
}
\usepackage{pifont}
\usepackage{multirow}
\usepackage{amsmath}
\usepackage{rotating}
\usepackage[dvipsnames]{xcolor}
\usepackage{comment}
\usepackage{amssymb}

\usepackage{url}
\usepackage{hyperref}
\usepackage{array} %

\newcommand{\mb}[1]{\boldsymbol{#1}}
\newcommand{\mseq}[1]{\mb{#1}}
\begin{document}
\sloppy

\title{DANIEL: A fast Document Attention Network for Information Extraction and Labelling of handwritten documents}

\author{Thomas Constum
    \and
    Pierrick Tranouez
    \and
    Thierry Paquet
}

\institute{T. Constum \at
              LITIS, University of Rouen Normandie, Rouen, France \\
              \email{thomas.constum1@univ-rouen.fr} %
           \and
           P. Tranouez \at
              LITIS, University of Rouen Normandie, Rouen, France \\
              \email{pierrick.tranouez@univ-rouen.fr} %
           \and
           T. Paquet \at
              LITIS, University of Rouen Normandie, Rouen, France \\
              \email{thierry.paquet@univ-rouen.fr} %
}

\date{Received: date / Accepted: date}

\maketitle

\begin{abstract}
Information extraction from handwritten documents involves traditionally three distinct steps: Document Layout Analysis, Handwritten Text Recognition, and Named Entity Recognition. Recent approaches have attempted to integrate these steps into a single process using fully end-to-end architectures. Despite this, these integrated approaches have not yet matched the performance of language models, when applied to information extraction in plain text.
In this paper, we introduce DANIEL (Document Attention Network for Information Extraction and Labelling), a fully end-to-end architecture integrating a language model and designed for comprehensive handwritten document understanding.
DANIEL performs layout recognition, handwriting recognition, and named entity recognition on full-page documents.
Moreover, it can simultaneously learn across multiple languages, layouts, and tasks.
For named entity recognition, the ontology to be applied can be specified via the input prompt.
The architecture employs a convolutional encoder capable of processing images of any size without resizing, paired with an autoregressive decoder based on a transformer-based language model.
DANIEL achieves competitive results on four datasets, including a new state-of-the-art performance on RIMES 2009 and M-POPP for Handwriting Text Recognition, and IAM NER for Named Entity Recognition.
Furthermore, DANIEL is much faster than existing approaches.

We provide the source code and the weights of the trained models at \url{https://github.com/Shulk97/daniel}.


\end{abstract}

\section{Introduction}
\label{sec:intro}

The challenge of understanding handwritten documents persists as a significant barrier in historical research, leaving a treasure trove of documents rich in invaluable information largely untapped.
Indeed, it would be too costly and time-consuming to analyze these documents by hand entirely.
This is why Handwritten Document Recognition (HDR) has emerged as a means of automatically extracting transcriptions from handwritten documents.

Traditionally, HDR relies first on Document Layout Analysis (DLA), in order to detect the text lines, then Handwritten Text
Recognition (HTR) is conducted to generate
transcriptions.
Yet, recent advancements suggest a promising direction: the integration of HTR and DLA through innovative architectures capable of interpreting paragraphs and entire documents. One such leading architecture in HDR is DAN \cite{coquenet_dan}, an encoder-decoder framework recognized for setting the current benchmark in the field.

However, merely extracting textual content indiscriminately from handwritten documents seldom aligns with the overarching objective of automatically extracting key information from the documents.
This task, known as Information Extraction (IE), necessitates a dedicated step following text recognition. This step involves locating and labeling the relevant information within the entire document transcript.
Previously treated as a separate task, IE has recently been integrated into fully end-to-end architectures combining HTR, layout recognition, and IE using a variant of DAN \cite{constum2024endtoend}.
This evolution heralds the emergence of Handwritten Document Understanding (HDU) as a distinct and burgeoning research domain.

Very close to HDU, Visual Document Understanding (VDU), focused on deriving actionable insights from
printed business documents, like invoices or receipts, but also on interpreting complex structures like tables, schemes, and images. This field of research has embraced similar methodological evolutions.
Transitioning from reliance on OCR-dependent approaches to embracing end-to-end architectures mirroring the encoder-decoder architecture exemplified by DAN, VDU illustrates a convergence with HDU towards a unified architecture paradigm.

This trend shows a pivotal shift towards using universal encoder-decoder architectures that can cater to a broad spectrum of document categories.
Whether through a Convolutional Neural Network (CNN) or a Vision Transformer (ViT) \cite{vit}, the encoder's versatility, coupled with an autoregressive transformer decoder, showcases diverse capabilities.
The decoders' distinction mainly lies in their scale, with large-scale 
models like Donut \cite{kim_ocr-free_2022} or Dessurt \cite{davis_end--end_2022} boasting internal dimensions ranging between 768 and 1024, in contrast to lighter models such as DAN, with a 256 decoder dimension.
This diversity extends to the output of these models, encompassing 
for instance
joint layout and text recognition, document classification or visual question answering.

A significant merit of larger architectures over DAN is their advanced language modeling capabilities, achieved via transfer learning or model distillation from pre-trained language models.
Indeed, the integration of pre-trained language models in the decoder side of the architecture improves the overall capability of the model not only for the text recognition task but most importantly for labeling the textual content providing a nuanced understanding of language, which is
crucial for excelling in tasks like IE.
Conversely, DAN's strength lies in its convolutional encoder, capable of accommodating any image size or aspect ratio without resizing, a critical advantage for handwritten document recognition, given the variability in document and character sizes.

For these reasons, we propose to combine a convolutional encoder for versatile image input, with a pretrained transformer-based language model used as a
decoder for refined language modeling.
The proposed approach also benefits from model distillation techniques using a DeBERTa v3 model \cite{he_debertav3:_2023} trained on named entity recognition (NER) to enhance its language understanding capabilities.

Furthermore,
transformers' reliance on vast training datasets is a notable challenge, especially when the goal is to accurately model the nuances of handwritten documents in terms of layout and handwriting styles.
A prevalent strategy to circumvent this obstacle involves the use of synthetic data, which, however, has its own limitations.
For instance, DAN employs a relatively narrow selection of fonts and a constrained textual corpus based on the training labels for its synthetic data.
This approach is relevant for small-scale models, as only a few fonts are necessary to pre-train the visual encoder.
Moreover, this choice of textual corpus ensures that language modeling closely mirrors the target data. However, using a small set of fonts and a limited textual corpus 
can lead to model overfitting in more expansive frameworks.
Indeed, a large encoder-decoder model might show its transformer's language modeling capabilities to depend exclusively on textual cues from its self-attention layer, rather than visual input to make its predictions. In extreme scenarios, this reliance solely on linguistic patterns can result in the generation of sentences that are grammatically correct but have no link with the input image.

In contrast, larger VDU models like Donut \cite{kim_ocr-free_2022}, Dessurt \cite{davis_end--end_2022}, or Pix2Struct \cite{lee_pix2struct:_2023} incorporate synthetic data boasting a wider array of visual and linguistic elements but often lack a sufficient representation of handwritten fonts.
In fact, these models are designed primarily to process printed commercial documents.
Some works utilizing handwritten fonts in their synthetic data have been proposed recently, but they are limited to line image generation \cite{barrere_training_2024,kang_pay_2022}.

To address these shortcomings and enhance the versatility of the model, we 
introduce a sophisticated synthetic
document page generator. This generator is equipped with an expansive library of 600 handwritten fonts and
includes texts in English, French, and German, aiming to significantly broaden the model's applicability and learning potential.

Since its introduction in 2022, DAN is a benchmark for full-page handwriting recognition. Nonetheless, its relatively slow inference speed inherent to its autoregressive, character-level prediction process has been a notable limitation.
In response, the Faster-DAN \cite{coquenet_faster_2023} variant was proposed, offering enhanced inference speeds thanks to the simultaneous recognition of all text lines. However, this gain in speed came at the expense of recognition accuracy degradation.
Our proposal navigates these challenges adeptly by combining a subword-scale prediction mechanism and an optimized implementation to get a remarkable speed, while performing better or equally with other existing methods.

In this work, %
we introduce a Document Attention Network for Information Extraction and Labelling (DANIEL), a groundbreaking end-to-end architecture for handwritten document understanding.

The main contributions of our work are summarized as follows:
\begin{itemize}
    \item We propose the Document Attention Network for Information Extraction and Labelling (DANIEL), a fully end-to-end architecture performing Layout Analysis, Handwritten Text Recognition and Named Entity Recognition on full-page documents.
    \item With its fully convolutional encoder, DANIEL can handle document images of any size and any aspect ratio without resizing.  
    \item We design a new pre-training method that trains DANIEL simultaneously on multiple layouts, languages, and tasks. 
    Unlike VDU pre-training methods, this method is specially designed for processing handwritten documents including documents with complex layout, and is suitable for large models which are prone to overfitting, thanks to the integration of a rich diversity of visual and linguistic aspects.
    \item The proposed network achieves a new state-of-the-art performance for HTR on RIMES 2009 and M-POPP and competitive results on IAM and READ 2016.
    \item We demonstrate that applying model distillation from a large language model enables DANIEL to achieve a new state-of-the-art performance for NER on IAM NER even outperforming sequential methods. DANIEL also achieves state-of-the-art results on M-POPP NER.
    \item DANIEL brings a new state of the art in terms of speed on every dataset it was evaluated compared to existing full-page text recognition and named entity recognition architectures.
    \item We provide the source code, the synthetic data generators, and the weights of the trained models.
\end{itemize}

This paper is organized as follows. 
Section 2 reviews related work in Handwritten Text Recognition, Named Entity Recognition, and Visual Document Understanding.
Section 3 details our architecture.
Section 4 describes the real datasets and the synthetic datasets that we designed to train the model.
Section 5 outlines the pretraining strategy.
Section 6 compares various fine-tuning strategies, presenting the results in terms of Handwritten Text Recognition, Named Entity Recognition, and inference speed.

\section{Related Works}
The convergence of deep learning techniques in handwriting text recognition, named entity recognition, and visual document understanding has transformed the landscape of document analysis. This section explores the evolution and integration of these domains, highlighting how advancements have enabled comprehensive and automated interpretations of both the textual and visual content of handwritten documents, enhancing the computer ability to extract and contextualize information efficiently.

\subsection{Handwritten Text Recognition}\label{related-works-htr}

In the domain of handwritten text recognition, the transition towards comprehensive document analysis has marked a pivotal shift, moving beyond the traditional confines of isolated line or word recognition. Earlier strategies necessitated the pre-segmentation of documents into more manageable units like lines or words and employed a variety of methods including MD-LSTM \cite{voigtlaender_handwriting_2016}, CRNN \cite{puigcerver_are_2017,wigington_data_2017} (a combination of CNN and BLSTM networks) or solely CNN-based approaches \cite{coquenet_have_2019,coquenet_recurrence-free_2020}. These techniques primarily leveraged Connectionist Temporal Classification \cite{CTC} (CTC) to adeptly navigate the alignment challenges intrinsic to handwriting variability during the training process.

While CTC was initially conceived for addressing one-dimensional alignment problems, its application has been ingeniously extended to the recognition of text across paragraphs. This extension involved either reinterpreting the inherently two-dimensional nature of paragraphs into a linear format \cite{coquenet_span:_2021,yousef_origaminet} or employing a recurrent approach to implicitly segment lines prior to applying CTC \cite{bluche_joint_2016,coquenet_end}, thus facilitating the recognition of continuously written text.

Simultaneously, with the evolution of attention-based models like transformers solving the alignment problem, the cross-entropy loss was demonstrated to be sufficient to account for character sequence recognition.

While early methods were using LSTM \cite{rostock-2019} or MD-LSTM \cite{bluche2016scan} models, recent ones are based on transformers \cite{kang_pay_2022}.
The incorporation of attention mechanisms enables these architectures to adeptly handle text recognition tasks not just at the line level \cite{rostock-2021,wick_rescoring_2022,kang_pay_2022,li_trocr:_2022,barrere_light_2022,barrere_training_2024} but also over entire paragraphs \cite{rouhou_transformer-based_2022,bluche2016scan} and in the case of transformer-based methods also over complete pages \cite{singh_full_2021,coquenet_dan}.
Most of the transformer-based approches rely on using synthetic data during training to prevent  overfitting, an inherent phenomenon to these architectures.

However, the sequential nature of prediction in transformer models often results in slow inference speeds when compared to methodologies that predict every line simultaneously.
To counteract the latency in inference speeds inherent to transformer-based architectures, innovations such as the Faster-DAN \cite{coquenet_faster_2023} have been introduced. This variant of the DAN \cite{coquenet_dan} enhances processing speed by parallelizing line predictions, although this increase in efficiency may sometimes come at the expense of recognition accuracy.
Similar methods to improve the inference speed of autoregressive models exist in other fields such as machine translation \cite{2019maskpredict} or speech recognition \cite{tian2020spiketriggered}.

\subsection{Named Entity Recognition}

Information Extraction (IE) in digitized documents generally unfolds in a three-step pipeline: document image segmentation, text recognition,
 and Named Entity Recognition (NER) on the transcription.
 A notable challenge within this framework is the cascading effect of errors: mistakes at any stage adversely affect the accuracy of subsequent steps. This sequential methodology not only demands annotated datasets and stage-specific training, but adjustments in any one phase may require retraining the downstream processes. Moreover, processing extensive datasets in this manner generates a plethora of intermediate files, particularly during segmentation, leading to substantial storage demands.

As detailed in section \ref{related-works-htr}, methodologies for Handwritten Text Recognition (HTR) have been developed to operate at the page level.
When integrated with NER techniques, such as those utilizing the BERT \cite{devlin_bert:_2019} language model, a comprehensive IE pipeline emerges.

Concurrently, significant strides have been made in executing HTR and IE in an end-to-end manner, both at line level \cite{carbonell_joint_2018,wigington_multi-label_2019,boros_comparison_2020,tarride_comparative_2022} or at page level by employing Feature Pyramid Networks for generating word bounding boxes \cite{carbonell_neural_2020}. These strategies, known as combined or integrated approaches, offer a promising alternative to traditional sequential methods.

Similarly, recent advancements have introduced a segmentation-free architecture that seamlessly integrates DLA, HTR, and NER capabilities.
Notably, several innovations leverage
an encoder-decoder framework \cite{rouhou_transformer-based_2022,tarride_key-value_2023,constum2024endtoend}, which combines a convolutional encoder with a transformer decoder. Distinctively, another approach adopts a fully transformer-based architecture, uniquely structured into three branches \cite{davis_end--end_2022}. 

In the literature, very few datasets can serve for benchmarking both handwriting recognition and NER.
The first dataset to allow such a benchmark was Esposalles \cite{romero_esposalles_2013} which comprises handwritten marriage records in ancient Catalan.
It is crucial to acknowledge that, given the advancements in model performance, the Esposalles dataset's relevance for current model evaluation has diminished, with top method \cite{constum2024endtoend} now achieving a 96.84\% IEEHR metric \cite{fornes_icdar2017_2017}.

Recently, the IAM dataset \cite{marti_iam-database} was augmented with Named Entity annotations providing a more challenging dataset. The IAM dataset is an HTR dataset that consists of English handwritten paragraphs with sentences extracted from the LOB corpus and the IAM NER dataset is an augmented version that includes named entity annotation following the OntoNotes ontology \cite{pradhan_towards_2013}.
Very recently, M-POPP \cite{constum2024endtoend}, a dataset containing French handwritten marriage records was proposed as a third challenging benchmark. 
Another dataset for IE from handwritten documents is SIMARA \cite{tarride_simara:_2023}.
Nonetheless, this dataset is not a NER dataset per se, as it does not comprise actual sentences, which makes it more of a key-value extraction dataset.

The debate between integrated (end-to-end) and sequential IE methods remains unresolved. Integrated approaches, as demonstrated, can enhance both HTR and IE performance by expanding contextual understanding through concatenated line predictions \cite{carbonell_joint_2018}. However, limited evidence also suggests that sequential methods, particularly those employing advanced language models like RoBERTa \cite{liu_roberta:_2019} pre-trained for NER, could surpass integrated models in efficacy \cite{tuselmann_are_2021} when working on segmented word images, although the end-to-end methods described here skip the transcription part altogether, which is not our objective.

\subsection{Visual Document Understanding}

Visual Document Understanding (VDU) has evolved to cover a wide spectrum of tasks beyond simple text recognition. These tasks include for instance document classification, visual question answering, and form understanding, focusing on extracting pivotal information from semi-structured documents, predominantly those that are printed. For these tasks, the research community has increasingly leaned towards employing pre-trained transformers for their versatility, sparked by breakthroughs in language models like BERT \cite{devlin_bert:_2019}. Among the pioneering models, Layout-LM \cite{xu_layoutlm:_2020} stands out as the first to synergistically model both text and layout, leveraging the capabilities of BERT. It integrates OCR-generated text and bounding box coordinates with patch images derived from these boxes, marking a significant advancement in the field.

Subsequent innovations have introduced variants and improvements to Layout-LM, such as BROS \cite{hong_bros:_2022}, which eschews visual features in favor of a relative spatial encoding and zone masking training strategy, and Layout-LM v2 \cite{xu_layoutlmv2:_2022}, enriching the model with visual tokens and supplementary pre-training tasks aimed at refining text-image alignment.
The introduction of LayoutLMv3 \cite{huang_layoutlmv3:_2022} marks another leap in this evolution by eliminating the need for convolutional neural networks (CNNs) for image feature extraction, using a unified text and image masking technique during pre-training instead. This approach simplifies the architecture and enhances efficiency across various VDU tasks. By incorporating a word-patch alignment objective, LayoutLMv3 finely tunes the model's understanding of the complex interplay between textual and visual document components.

Traditionally, these methods have adopted sequence labeling approaches, grounded in encoder-only transformers. However, the exploration of generative approaches, such as the one mentioned in TILT \cite{powalski_going_2021}, addresses the limitations inherent in extractive methods. This is particularly useful in scenarios like Visual Question Answering (VQA) where the answer may not be directly retrievable from the image, showcasing the adaptability and innovative progress in the field of VDU.

Recent methodologies have mirrored those in HTR and NER, moving towards end-to-end solutions by eliminating reliance on classical Optical Character Recognition (OCR) technology. 
This shift is exemplified by innovative image-to-text models like Dessurt \cite{davis_end--end_2022} and Donut \cite{kim_ocr-free_2022}, which utilize transformer-based designs to deliver a thorough understanding of both the document's layout and its content without using segmentation annotation.
These architectures are capable of performing various tasks which are indicated to the model via the input prompt. This input prompt can be a simple start token or a question as is the case for VQA.
Donut, employing an encoder-decoder structure, integrates a SWIN \cite{liu_swin_2021} encoder with an mBART \cite{liu_multilingual_2020} decoder, both of which are pre-trained to enhance the model's efficiency. Initially, Donut undergoes pre-training through a simplified OCR task on synthetic data, setting the stage for further specialization in areas such as Document Classification, Visual Question Answering (VQA), or Document Information Extraction through fine-tuning.
Dessurt takes a similar path but differentiates itself by incorporating
an array of real and synthetic datasets specialized for different downstream tasks as well as pre-training tasks such as masked language modeling applied to images.
This comprehensive pre-training regimen allows Dessurt to demonstrate its versatility across nine different dataset-task pairings, including HTR, underscoring its adaptability to various document understanding challenges.

Adding to the landscape, Pix2Struct \cite{lee_pix2struct:_2023} introduces an innovative pre-training task focused on screenshot parsing. This task entails the transformation of masked web page screenshots into simplified HTML structures, marking a notable step forward in the model's ability to parse and understand complex web-based documents.
Pix2Struct innovatively integrates language prompts (e.g., questions) directly onto the input image during finetuning, treating all inputs through a single modality. 
This helps maintain both visual and textual context together, which is crucial for tasks involving visually-situated language.
Recently, Nougat \cite{blecher_nougat:2023} introduced an advanced document understanding framework based on the Donut model and specifically tailored for processing academic texts. This framework converts PDFs into a structured markup language, facilitating the handling of complex elements, such as mathematical formulas, which are prevalent in many domains.

Recent methods predominantly utilize transformer architectures following the encoder-decoder paradigm. These approaches leverage synthetic data for both printed and handwritten documents. The primary distinction among these methods lies not in the architectures themselves, but in their application. 
Indeed, these models demonstrate versatility in performing a wide range of tasks.
Regarding handwritten documents,
some methods have focused exclusively on text and layout understanding \cite{coquenet_dan}, while others have concentrated solely on handwriting recognition and NER \cite{rouhou_transformer-based_2022}.
A method \cite{constum2024endtoend} has been applied to layout understanding, handwriting recognition, and named entity extraction of handwritten documents, but it does not incorporate a pre-trained language model.
However, the integration of language models through transfer learning or model distillation is a key advantage of large Document Understanding architectures, as it significantly enhances their language modeling capabilities, especially for NLP tasks such as NER.
The efficacy of such architectures has been evaluated on commercial documents, but not on handwritten documents.
The key difference between these two types of documents is that commercial documents contain pre-structured information, whereas handwritten documents require locating the information within the text.
There is a lack of research on best practices for training models that incorporate language models for handwritten documents especially to avoid overfitting on training data.
Therefore, this article presents the first study of an encoder-decoder architecture applied to handwritten documents, combining layout recognition, handwriting recognition, and named entity extraction, and incorporating a pre-trained language model.

\begin{figure*}[t!h]
    \centering
    \includegraphics[width=0.95\linewidth]{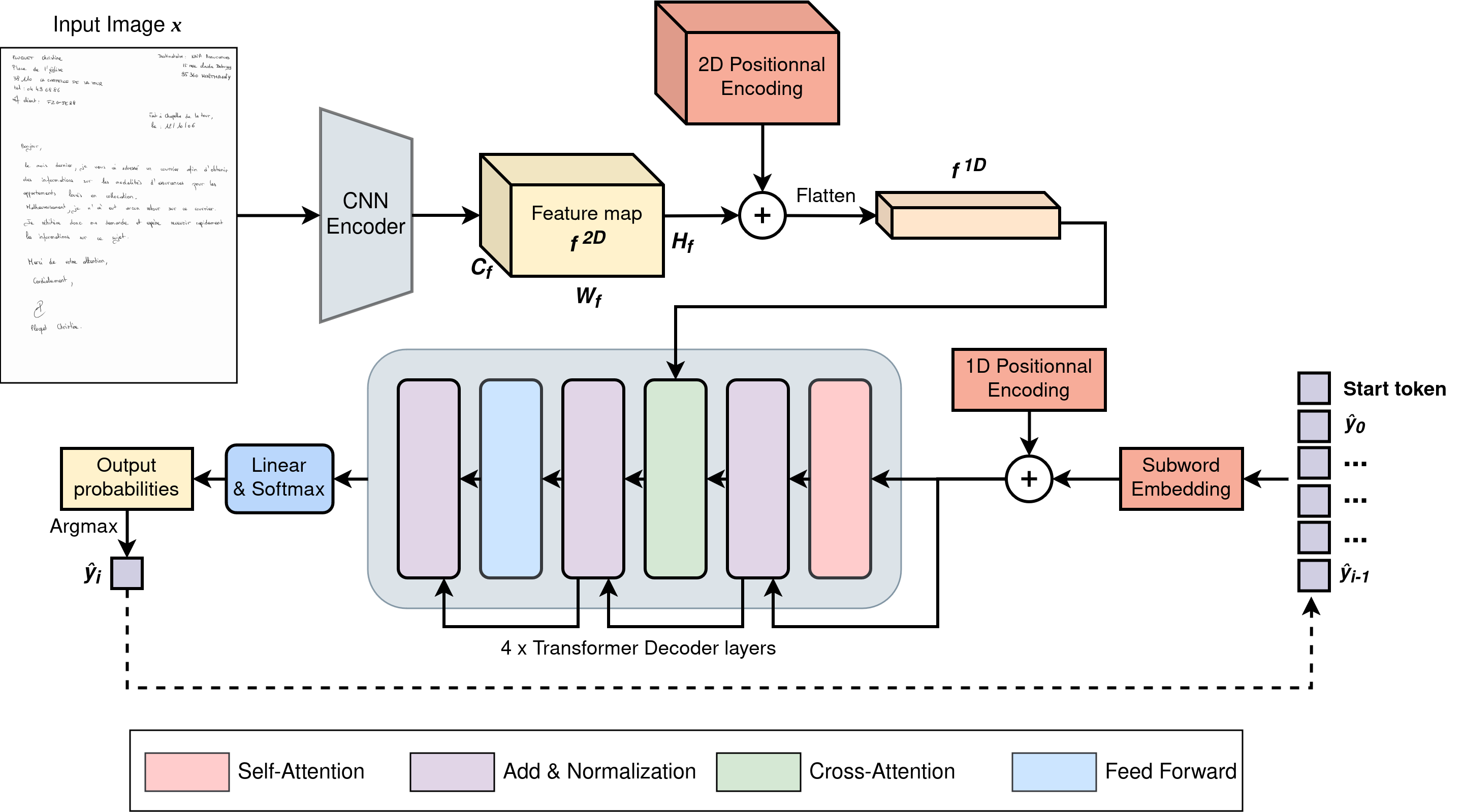}
    \caption{Diagram of the DANIEL architecture. DANIEL comprises a convolutional encoder that extracts a 2D feature map $\mb{f}_\mathrm{2D}$ and a transformer-based decoder for the sequential prediction of subword, layout, and named entity tokens $\hat{\mseq{y}}_i$.
    Prediction begins with a
    start
    token, provided as a query, which depends on the task at hand. At each step $i$, the model predicts the $i^{th}$ token, which is then appended to the query for the subsequent prediction step.
    }
    \label{fig:diagramme-archi}
\end{figure*}

\section{DANIEL architecture}
DANIEL is an end-to-end
architecture that is able to perform full page HTR and
NER on handwritten documents. 

This architecture is
based on a 
image encoder and a language decoder
and
takes as input 
two types of information: an image and a start token as a query indicating which task to perform to the model. 
It outputs a sequence of tokens depending on the target task. 
The model uses subword tokens to encode the textual content but also layout tokens as well as named entity tokens.
The architecture is shown in Fig \ref{fig:diagramme-archi}.

\subsection{Encoder}
\label{presentation-encoder}
For the encoder, we use an up-scaled version of the DAN encoder. 
Contrary to most of the VDU architectures such as Donut \cite{kim_ocr-free_2022} or Dessurt \cite{davis_end--end_2022}, the encoder of DANIEL is not transformer-based, it
 pivots to a convolutional architecture, thereby enabling flexibility in handling various input sizes and aspect ratios. This flexibility is crucial for avoiding alterations in the appearance of characters in the images.
In contrast, reference architectures like Donut and Dessurt are constrained by fixed input image sizes of $2560\times1920$ and $1152\times768$ pixels, respectively. Similarly, Pix2Struct is limited to processing a fixed number of image patches, specifically $4096$, which restricts its flexibility compared to DANIEL.

Given the wide variation in dimensions and aspect ratios among handwritten documents, standardizing image sizes through resizing can lead to significant discrepancies in character representation. Specifically, documents of the same size can exhibit vastly differing character sizes, highlighting the limitations of a one-size-fits-all resizing approach.

We designed the encoder of DANIEL by upscaling the original DAN encoder,  doubling the embedding size of the initial four blocks. Subsequently, the embedding sizes for the following five blocks are increased from 256 to 512, and the final block is configured to 1024. 
Table \ref{details-encodeur} gives 
information about the number of output channels for each block and the corresponding block type. The two types of layer blocks used are Convolutional Block (CB) and Depthwise Separable Convolutional Block (DSCB). More information on these blocks can be found in \cite{coquenet_end}.
For more details on the choice of the encoder see \ref{encoder-comparison}.

The encoder transforms the input document image, represented as
$x\in\mathbb{R}^{H\times W\times C}$
into a feature map \( f^{2D} \), where \( f^{2D} \in \mathbb{R}^{H_f \times W_f \times C_f} \). Here, \( H \), \( W \), and \( C \) denote the input image's height, width, and number of channels, respectively, with \( H_f = \frac{H}{32} \), \( W_f = \frac{W}{8} \), and \( C_f = 1024 \). We specifically consider grayscale images, hence we set $C=1$.

Subsequently, the feature map $f^{2D}$ is summed with
the
2D positional encoding.
The augmented feature map is then flattened to produce a 1D feature map $f^{1D}$, where \( f^{1D} \in \mathbb{R}^{(H_f\times W_f) \times C_f} \). This 1D feature map $f^{1D}$ is subsequently fed into the decoder for further processing.

\renewcommand{\arraystretch}{1.25}
\begin{table}[ht]
    \centering
    \resizebox{\columnwidth}{!}{
    \begin{tabular}{c|c|c}
        \hline
         & \textbf{Encoder of DAN} & \textbf{Encoder of DANIEL} \\\hline
        output size & (block type, \# output channels) & (block type, \# output channels) \\
        \hline
        $H \times W$  & $(CB, 16) \times 1$ & $(CB, 32) \times 1$ \\
        \hline
        $\frac{H}{2}\times\frac{W}{2}$  &
        $(CB, 32) \times 1$ &
        $(CB, 64) \times 1$ \\
        \hline
        $\frac{H}{4}\times\frac{W}{4}$  &
        $(CB, 64) \times 1$ &
        $(CB, 128) \times 1$ \\
        \hline
        $\frac{H}{8}\times\frac{W}{8}$  &
        $(CB, 128) \times 1$ &
        $(CB, 256) \times 1$ \\
        \hline
        $\frac{H}{16}\times\frac{W}{8}$  &
        $(CB, 128) \times 1$ &
        $(CB, 512) \times 1$ \\
        \hline
          &
        $(CB, 128) \times 1$ &
        $(CB, 512) \times 1$ \\
        $\frac{H}{32}\times\frac{W}{8}$ &
        $(DSCB, 128) \times 3$ &
        $(DSCB, 512) \times 3$ \\
         &
        $(DSCB, 256) \times 1$ &
        $(DSCB, 1024) \times 1$ \\
        \hline
        \# params & $1.7 \times 10^6$ & $20.0 \times 10^6$ \\
        \hline
    \end{tabular}
}
\caption{Comparative table of the encoders of DAN and DANIEL. CB stands for Convolutional Block and DSCB for Depthwise Separable Convolutional Block.}
\label{details-encodeur}
\end{table}
\renewcommand{\arraystretch}{1.0}

\subsection{Decoder}

As a decoder, we employ the 4 first transformer blocks from the mBART decoder \cite{liu_multilingual_2020}. 
This choice capitalizes on its well-established 
language modeling
capabilities.
We adopt Donut's tokenizer, which utilizes the SentencePiece tokenization methodology \cite{kudo_sentencepiece:_2018}. However, we refine its vocabulary as detailed in  \ref{vocab}.
It is crucial to emphasize that our focus is not on developing a character recognizer, but rather a subword recognizer.
Consequently, the model is required to distinguish among tens of thousands of tokens, a significant escalation from traditional systems that typically discriminate among only a few hundred characters.

The decoder processes the feature map $f^{1D}$ alongside a task-specific start token to produce a token sequence $(\hat{y}_i)_{i=1}^{m}$, with each $\hat{y}_i\in\mathbb{R}^v$ representing a one-hot vector corresponding to the $i$-th token. Here, $v$ denotes the vocabulary size, while $m$ is set as a hyperparameter to determine the maximum length of the generated sequence.
The final predicted sequence is decoded into text via the greedy decoding method.

\section{Datasets}

\subsection{Real datasets}
To showcase the versatility and robust capabilities of our model, we conducted evaluations across multiple datasets, focusing on HTR and the combined application of HTR and NER, which we will refer to as HTR+NER.
Characteristics of the datasets and the dataset splits are provided in Tables \ref{tab:dataset-details} and  \ref{tab:tableau-splits} respectively.

\begin{table}[ht]
\centering
\resizebox{\columnwidth}{!}{%
\begin{tabular}{ccccc}
\hline
Dataset & \begin{tabular}[c]{@{}c@{}}\# entity\\ categories\end{tabular} & \begin{tabular}[c]{@{}c@{}}\# unique\\ words\end{tabular} & \begin{tabular}[c]{@{}c@{}}\# layout\\ tokens\end{tabular} & \begin{tabular}[c]{@{}c@{}}Dataset\\ level\end{tabular} \\
\hline
IAM & 18 & 14599 & 0 & paragraph \\
RIMES 2009 & 0 & 16789 & 14 & page \\
READ 2016 & 0 & 10023 & 10 & page \\
M-POPP & 118 & 11943 & 10 & page \\
\hline
\end{tabular}%
}
\caption{Details about the datasets.}
\label{tab:dataset-details}
\end{table}

\begin{table}[ht]
\centering
\begin{tabular}{cccc}
\hline
Dataset & Training & Validation & Test \\
\hline
IAM - split RWTH & 747 & 116 & 336 \\
IAM - split custom & 638 & 178 & 383 \\
RIMES 2009 & 1,050 & 100 & 100 \\
READ 2016 & 350 & 50 & 50 \\
M-POPP & 250 & 32 & 32 \\
\hline
\end{tabular}
\caption{Summary of dataset splits.}
\label{tab:tableau-splits}
\end{table}

\subsubsection{IAM}

The IAM dataset \cite{marti_iam-database} is a dataset written in English by 500 different authors which comprises modern documents sourced from the LOB corpus. 
This dataset is available at both line and paragraph levels. For the purposes of this article, we utilize the paragraph level, sometimes referred to as
IAM page,
since each document presents a single paragraph per page. 
An example of image from IAM page is shown in Fig 
\ref{fig:both-images-iam-rimes}.

The dataset images are stored in grayscale at a resolution of 300 dpi.
Initially developed solely for HTR, the IAM dataset was later annotated for NER by \cite{tuselmann_are_2021}, employing the OntoNotes v5 named entity ontology \cite{pradhan_towards_2013}.
This augmented version of IAM is called IAM NER.
This annotation is available in two formats: a comprehensive version with 18 named entity categories and a simplified version classifying entities into 6 categories only. 
In this article, we employ the 18-category version of IAM NER.

The dataset has multiple splits; in our research, we evaluate DANIEL on the RWTH split, tailored for HTR, and the \textit{custom} split designed for NER by \cite{tuselmann_are_2021}. 
Using only the RWTH split would be sub-optimal for NER and the same applies for the \textit{custom} split since no HTR performances have been reported in the literature on this split.

\subsubsection{RIMES}

\begin{figure}[h]
    \centering
    \fbox{\includegraphics[width=0.22\textwidth]{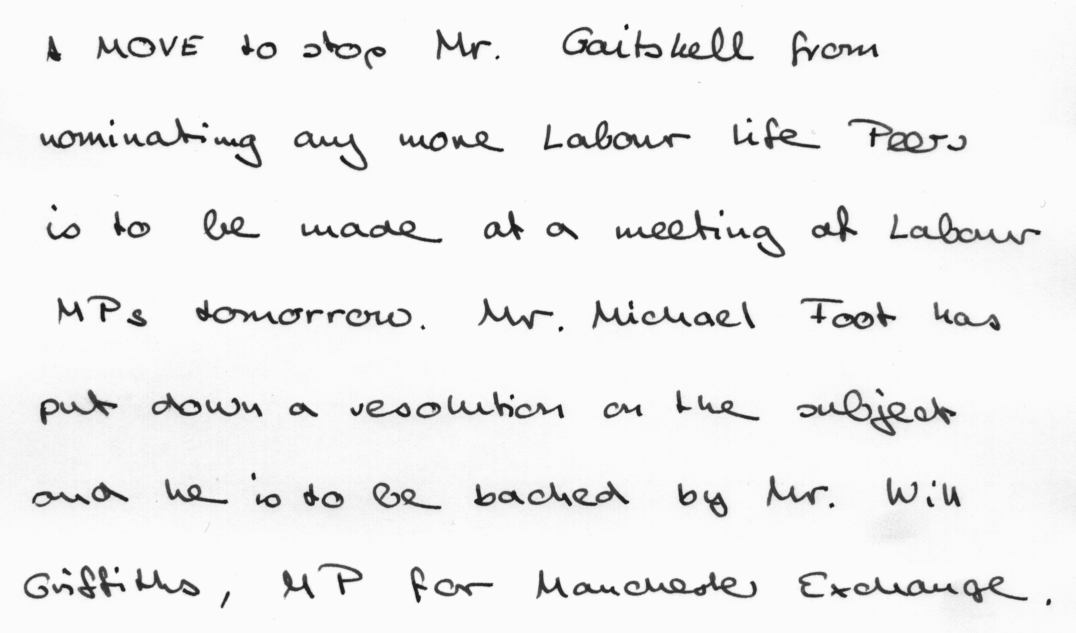}}
    \fbox{\includegraphics[width=0.22\textwidth]{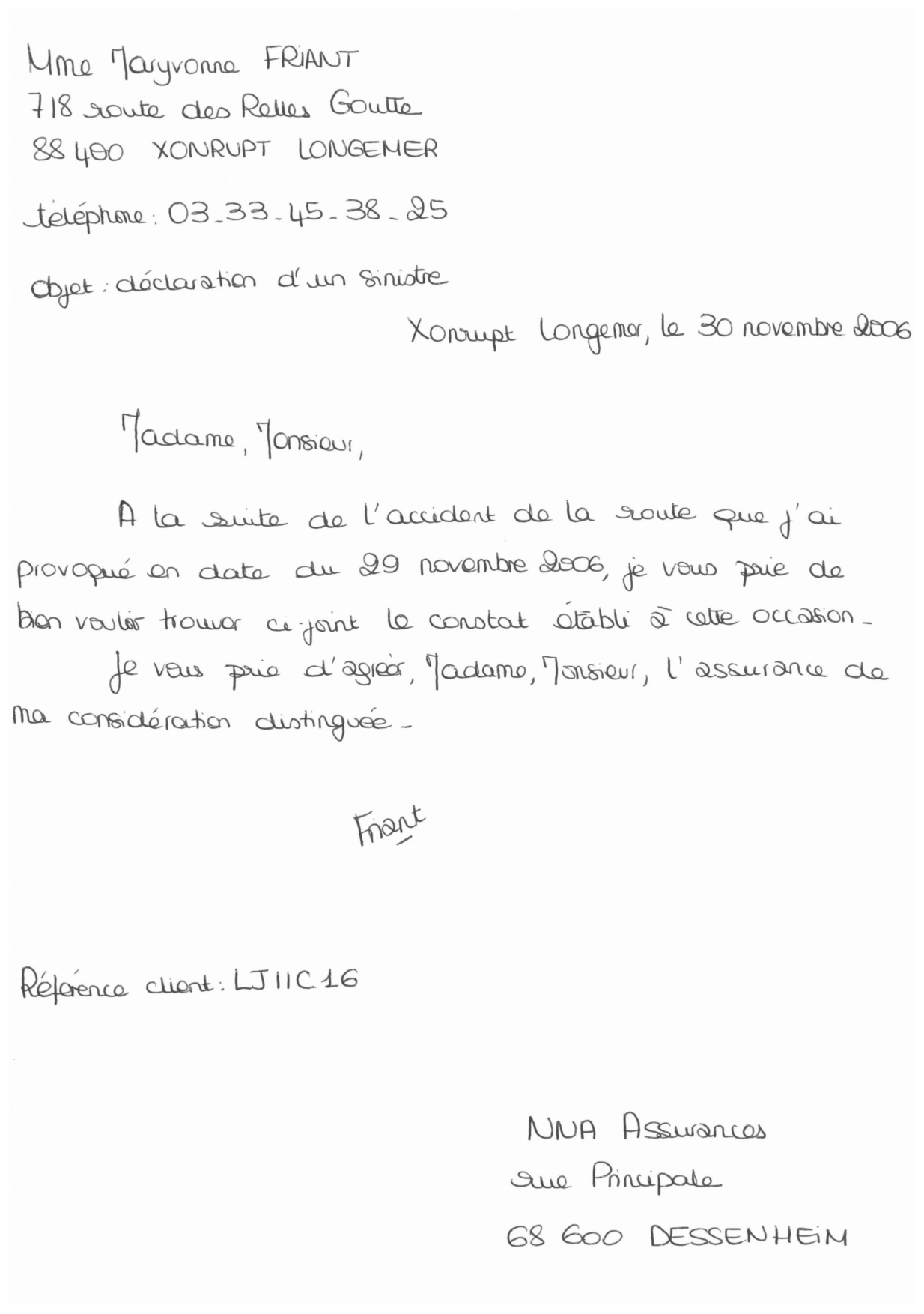}}
    \caption{Examples of images from the IAM (left) and RIMES 2009 (right) datasets.
    }
    \label{fig:both-images-iam-rimes}
\end{figure}

The RIMES dataset \cite{grosicki_results_2009}, a widely used collection of gray-scale images featuring French handwritten text, was produced in the context of mail writing. 
The dataset images are stored in grayscale at a resolution of 300 dpi.
Our evaluation focuses on the page-level variant of RIMES, known as RIMES 2009 whose images are available on Zenodo\footnote{\url{https://zenodo.org/records/10812725}}.
An example image from RIMES 2009 is shown in Fig
\ref{fig:both-images-iam-rimes}.
We use the same data partitioning, layout tokens, and reading order methodology as detailed in \cite{coquenet_dan}.
This dataset features a complex layout as each page is composed of a sequence of text blocks that can belong to seven different classes: sender, recipient, date \& location, subject, opening, body, and PS \& attachment.

\subsubsection{READ 2016}

The READ 2016 dataset \cite{sanchez_icfhr2016_2016}, a subset of the Ratsprotokolle collection from the READ project, was introduced during the ICFHR 2016 competition on handwritten text recognition. This dataset features Early Modern German handwriting. We utilize the page-level version, maintaining the same reading order and layout classes as specified in \cite{coquenet_dan}.
An example page image from READ 2016 is shown in Fig
\ref{fig:both-images-read-exopopp}.
Although the layout complexity of the READ 2016 dataset is marginally less than that of the RIMES 2009 dataset, it includes nested text blocks. The classes of text blocks in this dataset include page, page number, body, annotation, and section, the latter comprising a group of linked annotation(s) and body text.

\subsubsection{M-POPP}

\begin{figure}[h]
    \centering
    \includegraphics[width=0.22\textwidth]{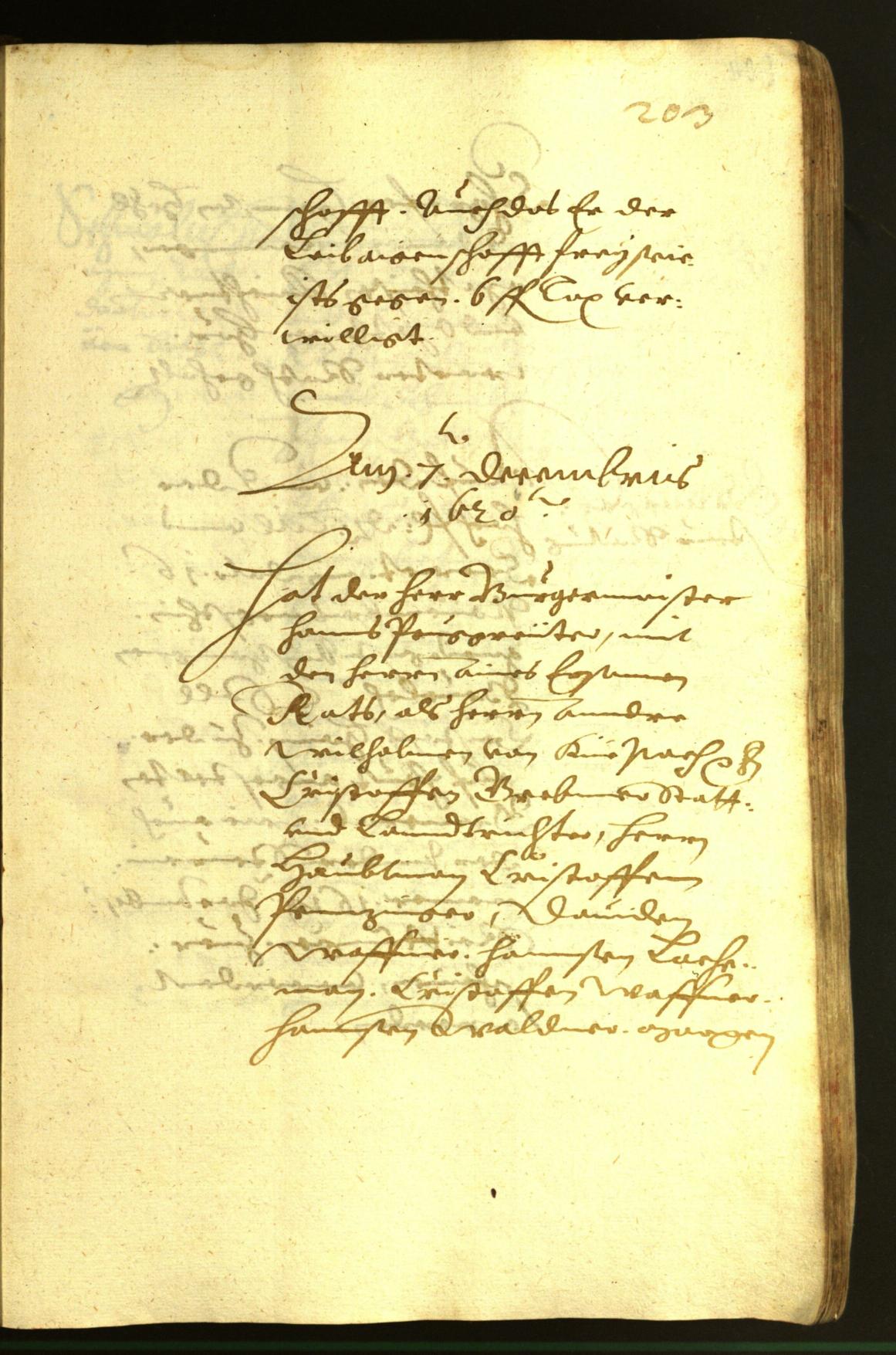}
    \includegraphics[width=0.22\textwidth]{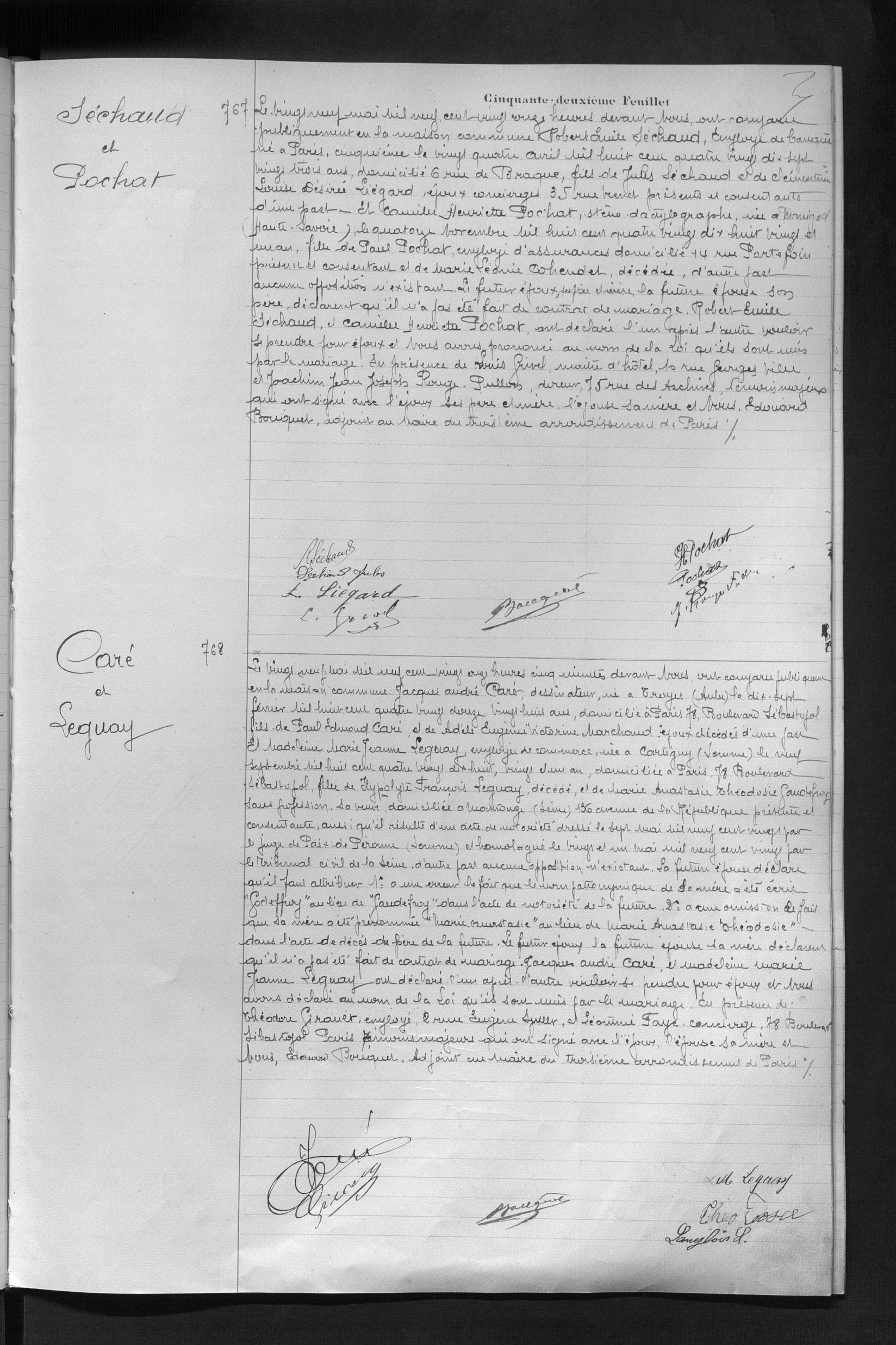}
    \caption{Examples of images from the READ 2016 (left) and M-POPP (right) datasets.}
    \label{fig:both-images-read-exopopp}
\end{figure}

The M-POPP dataset \cite{constum2024endtoend} is a dataset that is designed for full-page HTR and IE across both handwritten and printed documents. It was annotated as part of the Exo-POPP project, which focuses on extracting information from marriage records in Paris and its surrounding areas, covering the period from 1880 to 1940. 
In this article, we use the version 3 available on Zenodo\footnote{\url{https://zenodo.org/records/11296970}}.
An example of an image from M-POPP is shown in Fig
\ref{fig:both-images-read-exopopp}.
The layout of each page consists of three types of blocks: Blocks A and C are situated in the margins, with Block A containing the names of the married couple and Block C, which is optional, containing marginal notes. 
Block B, representing the main body of the text, is centrally located.
According to \cite{constum2024endtoend}, the dataset features over a hundred distinct writing styles.
For IE, each record can include up to 118 different information categories such as the occupation of the husband.
Our analysis concentrates on Blocks A and B to ensure comparability with \cite{constum2024endtoend} for both pure Handwritten Text Recognition (HTR) and the combined task of HTR and Information Extraction (HTR+IE).
Given that our study focuses on handwritten documents, we exclude the printed portion of the M-POPP dataset from our evaluation. 
For clarity, in the remainder of this article, \textit{M-POPP} refers to the dataset containing annotations solely for HTR, while \textit{M-POPP NER} refers to the dataset containing annotations for both HTR and NER.

\subsection{Synthetic data}
To the best of our knowledge, there is no large-scale datasets of handwritten documents currently available.
Consequently, it is imperative
to use synthetic data during the pretraining phase of the model.
The DANIEL architecture
is particularly prone to overfitting when trained on datasets of limited size due to its substantial number of parameters and its transformer-based decoder.
By using synthetic data, DANIEL not only learns to interpret handwritten text and understand the structure of various target documents but also
learns the language efficiently
which enhances its performance on NLP tasks, such as NER.
As a consequence, pre-training DANIEL on synthetic data requires not only generating text images with a high degree of writing variability but also generating text with as much variability as possible from the language perspective.
\subsubsection{HTR synthetic datasets}

Given the large scale of our model,
it is crucial to use a wide variety of fonts in order to avoid overfitting on the selected fonts.
To address this point, we employ a script referenced in \cite{davis_end--end_2022}, which scrapes the website 1001-fonts.com to obtain a significant collection of synthetic fonts.
We tailor the font selection for each language, considering that some languages include special characters not supported by all fonts.
Indeed, restricting ourselves to fonts that accommodate the characters of all languages would severely limit our options.
To align with our focus on handwritten document recognition, we predominantly use handwritten fonts in the generated datasets. However, we also incorporate a selection of printed fonts, which are more plentiful. The training process is designed to select fonts with an 80\%
probability
for handwritten styles and 20\% for printed styles, ensuring the model's
training
primarily
on
synthetic handwritten text. 

When generating synthetic data, the best method to ensure that language modeling closely mirrors the target data is to use text labels from the training sets of the target datasets.
However, the datasets used for training our model feature a relatively small number of samples, ranging from
250 pages in M-POPP
to 1050 pages in RIMES 2009. Such small datasets can be problematic for training large-scale transformer decoders like the DANIEL decoder, as they are prone to overfitting.
To counteract this problem, we supplement the training dataset with text from Wikipedia corpora in the targeted languages\footnote{For English and French, we utilize corpora from Huggingface: \url{https://huggingface.co/datasets/wikipedia}}\footnote{For German, we employed a corpus from the University of Leipzig, as the preprocessed German corpus from Huggingface was unavailable at the time of writing: \url{https://wortschatz.uni-leipzig.de/en/download/German}}, thereby enriching the diversity and quantity of the synthetic training data.
Moreover, to enhance the stability of the attention mechanism during line transitions, our method ensures continuity in language flow from one line to the next when generating the ground-truth text.

\begin{figure}[h]
    \centering
    \fbox{\includegraphics[width=5cm]{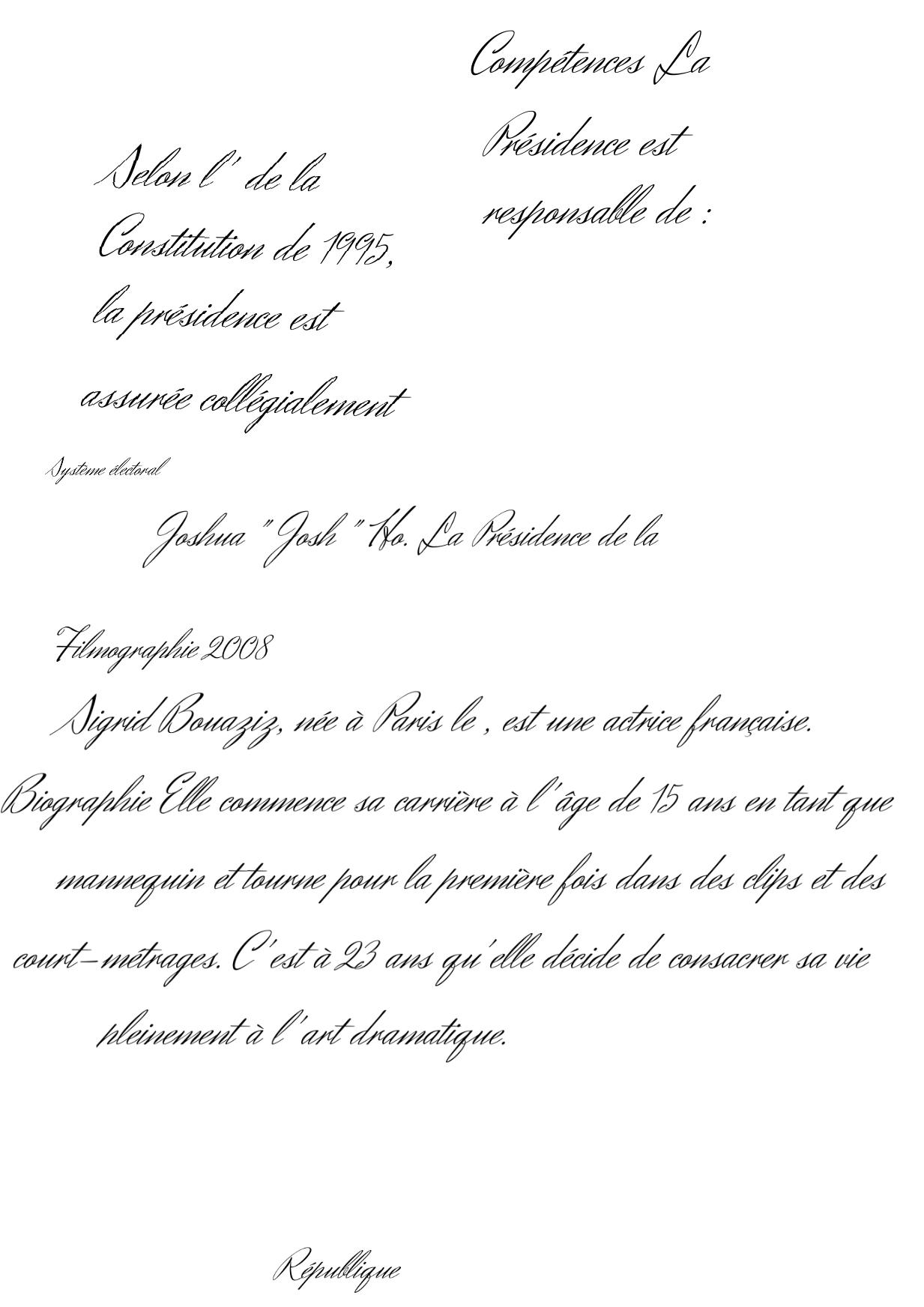}}
    \caption{Example of synthetic data for RIMES 2009.}
    \label{fig:exemple-rimes-synth}
\end{figure}

For each target dataset, we have customized a version of the synthetic dataset generator to accurately replicate the layout encountered within each dataset. For READ 2016 and RIMES 2009, 
we build on the synthetic data generators from \cite{coquenet_dan}. 
Specifically for RIMES 2009, modifications were required to adapt the DAN generator to our needs.
For each paragraph, we use text from Wikipedia, formatting it to mimic a RIMES text block. This is achieved by randomly selecting a real text block of the same type, then formatting the Wikipedia text to match the original
in terms of the number of lines and words.
For the M-POPP dataset, we employ the generator developed by \cite{constum2024endtoend}. 
Regarding the IAM dataset,
we designed a synthetic data generator
producing data consisting
of a single paragraph, structured with varying margins on the sides.
Figure \ref{fig:exemple-rimes-synth} shows an example of a synthetic page for RIMES 2009.

\subsubsection{HTR+NER synthetic datasets}
\paragraph{IAM NER}
For the IAM NER dataset, we use the synthetic data generator used for IAM and we add named entity annotations to the generated data.

Generating labels of synthetic data for HTR is straightforward using text corpora. However, creating synthetic data for named entity recognition (NER) poses more challenges due to the variability in the ontologies of named entities across different datasets and languages. For instance, since IAM NER utilizes the OntoNotes v5 ontology, it is necessary to use a corpus annotated with the same set of named entity categories.

To address this point, we employed a method akin to model distillation. This approach normally involves training a student model to mimic the output logits of a teacher model using a distillation loss \cite{hinton_distilling_2015}.
In our case, we use a language model trained on NER with the correct ontology to massively annotate text with named entities.
This annotated text is then utilized to generate synthetic data for IAM NER.
Training DANIEL on this data allows it to implicitly learn the representations taught by the teacher model, albeit through a different modality: the teacher model processes text input, while DANIEL processes image input. We use Deberta v3 Large \cite{he_debertav3:_2023}, a variant of BERT, as the teacher model. The used version was trained for NER following the OntoNotes v5 ontology\footnote{\url{https://huggingface.co/tner/deberta-v3-large-ontonotes5}}.
This model is used to annotate articles from the same Wikipedia corpus used by the synthetic data generator of IAM.

\paragraph{M-POPP NER}
Unlike the IAM dataset, 
which is based on very generic named entity categories,
M-POPP NER aligns more closely with an IE dataset, specifically targeting data from marriage certificates.
Creating a synthetic dataset for M-POPP NER would necessitate a substantial corpus of analogous marriage certificates and a language model specifically trained to annotate such data. To date, no such text corpus or language model is publicly accessible. 
Moreover, there are currently no large-scale datasets with sufficiently similar characteristics for effective named entity extraction.
Consequently, we did not utilize a synthetic dataset specifically for M-POPP NER in our study. Instead, we employed the synthetic HTR dataset of M-POPP.
This approach implies that for M-POPP NER, the pre-training phase on synthetic data primarily aids in learning handwriting recognition and layout patterns. Therefore, information extraction is exclusively learned through training on real data.

While it would have been feasible to train a Large Language Model (LLM) to generate and annotate documents akin to those in M-POPP, doing so would have involved extensive prompt engineering to guarantee the quality and diversity of the generated documents and annotations. Although this approach has not been pursued in this study, it represents a potential area for future investigation.

\section{Pretraining strategy}
Training transformer-based model is difficult, that is why existing methods in document understanding use synthetic data and pre-trained models to limit the amount of annotated data necessary for training.

The proposed pretraining strategy leverages synthetic data to ease the model's convergence.
Initially, the encoder is trained for the HTR task on synthetic lines.
In the subsequent phase, the model is trained to recognize multiple lines in the correct reading order, alongside the language modeling task, using synthetic documents. 
Finally, the model is fine-tuned on real data 
with different strategies described in section \ref{fine-tuning-strategies}.

\subsection{Encoder pre-training}
Since we are using a pre-trained language model as a decoder, it is necessary to
use a pre-trained encoder as well.
Indeed, connecting the pre-trained decoder with a randomly initalised encoder could damage the features learned by the decoder.

Hence, following the pre-training strategy of
\cite{coquenet_dan},
we first train the encoder alone
by creating a standalone line-level text recognition model from the encoder using the CTC loss at the character level.
This process entails the integration of an adaptive max pooling layer to collapse the vertical dimension, followed by the addition of a convolutional layer and the application of a softmax activation function.
We use the same hyper-parameters as in \cite{coquenet_dan}. 
Regarding the training data we use the same protocol
as for training the entire DANIEL model
except that the generation is made at the line level.
The encoder is pre-trained for 
40000 steps.

\subsection{Curriculum learning}
The next phase in the pre-training process
is devoted to training the entire model
on synthetic documents.
The encoder of the DANIEL is initialized with the weights of the pre-trained encoder from the previous pretraining step, while the decoder is initialized with the weights from the Donut decoder\footnote{\url{https://huggingface.co/naver-clova-ix/donut-base/tree/official}}.

During this phase, DANIEL is trained
on single-line documents and progressively increases the maximum number of lines per document to a specified upper limit, $l_{max}$, fixed for each target dataset. 
We crop the generated images just below the last written line, allowing the size of the images to expand proportionally with the number of lines.

\subsection{Teacher forcing}
Following the original Transformer \cite{vaswani_attention_2023}, we employ the teacher-forcing strategy during the training phase, where the
architecture
is trained to predict the next token in a sequence based on the preceding tokens using the ground truth data instead of relying on its own previous predictions, possibly erroneous.
To enhance DANIEL's resilience against prediction inaccuracies, we incorporate errors into the teacher forcing method.
Contrary to traditional text recognition models that operate at the character level, our model processes text at the subword level.
Given that the average length of subwords in the vocabulary is four characters,
randomly substituting one subword with another from the entire set of possible subwords could significantly distort the prediction process. This could potentially undermine the language model during training.

To mitigate this risk, we map each subword to a 
list of admissible replacement subword candidates 
that have a CER below a specific threshold, which varies depending on the length of the subword.
The longer the subword, the lower the CER threshold required, except for the longest subwords, which are infrequent in the vocabulary and consequently have fewer subwords with a small edit distance. 
We define a function $thresh_{cer}$ that assigns a CER threshold to each subword $x$, determining if another subword should be included in its list of close subwords. The function is defined as follows:

\[
thresh_{cer}(x) =
\begin{cases}
    1.5 & \text{if } length(x) \leq 2 \\
    0.7 & \text{if } length(x) = 3 \\
    0.5 & \text{if } 3 < length(x) < 9\\
    0.6 & \text{if } length(x) \geq 9 \\
\end{cases}
\]

\subsection{Training details}
Our methodology employs the data augmentation techniques previously utilized in \cite{coquenet_dan}.
For the RIMES 2009, READ 2016, and IAM datasets, we process images at a resolution of 150 dpi.
Similarly, for the M-POPP dataset, we maintain the image resolution from the version available on Zenodo\footnote{\url{https://zenodo.org/records/10980636}}, which is also 150 dpi.
It is worth noting that the DANIEL model is fully capable of operating on images at their original resolution. However, our experiments indicate that increasing the resolution beyond 150 dpi does not yield significant performance improvements, while substantially increasing the computational load and training time.
We normalize images to get a zero mean and unit variance. This standardization is based on the combined training sets across all datasets. 

For text recognition, we utilize a unique start token for all HTR datasets. In contrast, for NER, we use distinct start tokens for each dataset, specifically, one for IAM NER and another one for M-POPP NER. Additionally, our vocabulary also includes special tokens
for each named entity category.
Finally, for each layout category of each dataset, we define an opening and a closing tag.

Model training employs the Adam optimizer with a learning rate of $10^{-4}$ and a mini-batch size of 4, utilizing a single A100 GPU (80 GB).
For the teacher forcing, we use an error rate of 30\%.
We implement a curriculum learning strategy that adjusts the maximum number of lines per page ($l_{max}$) according to the unique requirements of each dataset: 40 for RIMES 2009, 15 for IAM and IAM NER, 30 for READ 2016, and 80 for M-POPP.
We employ a diverse selection of synthetic fonts tailored to the needs of each dataset: 5,998 fonts for IAM and IAM NER, 3,944 fonts for both RIMES 2009 and M-POPP, and 961 fonts for READ 2016. The variation in font selection across datasets is necessary to accommodate the unique characters specific to each language.

In this study, we investigate two types of pre-training: multilingual and monolingual. In multilingual pre-training, the model is simultaneously trained on several synthetic datasets while in monolingual pre-training, the model is pre-trained solely on the synthetic dataset corresponding to the target dataset.

For the multilingual pre-training applied to IAM, IAM NER, RIMES 2009, and READ 2016, the model is pre-trained concurrently on the respective synthetic datasets. When pre-training for the M-POPP dataset, the model is trained on all the synthetic datasets used in this study, including the synthetic M-POPP dataset.
The M-POPP dataset is handled separately due to its requirement for finer attention granularity compared to the other datasets mentioned.
To address this specificity, we implement the procedure described in \cite{constum2024endtoend}, which involves reducing the stride from 2 to 1 in block 5 of the encoder.
This adjustment results in the feature map at the encoder output being \( H_f = \frac{H}{16} \) instead of \( H_f = \frac{H}{32} \), following the notation in \ref{presentation-encoder}. While this modification increases training and inference times, it is exclusively applied to the M-POPP dataset to meet its specific requirements.

\section{Experiments and results}

\subsection{Setup and fine-tuning strategies}\label{fine-tuning-strategies}

In this article, we explore several fine-tuning strategies
using the weights obtained at the end of the corresponding pre-training process.
All strategies employ the same hyper-parameters used during pre-training, except the learning rate which is set to $10^{-5}$. The strategies we evaluate are as follows:

Strategy A (concurrent fine-tuning): The model is initialized with the weights of a model pre-trained using the multilingual setting and then fine-tuned concurrently on all real datasets.
Performance is assessed on the validation set of each dataset independently, retaining the weights that yield the best results for each respective validation set.

Strategy B (single dataset fine-tuning): The model is initialized with the weights of a model pre-trained using the multilingual setting and then fine-tuned on a single dataset.

Strategy C (sequential transfer fine-tuning): Initially, the model is fine-tuned using Strategy A. Once optimal performance is achieved for a specific target dataset, the best validation weights are used to initiate transfer learning on a new model. This new model is then fine-tuned solely on this target dataset, similar to Strategy B.

Strategy D (monolingual training): DANIEL is pre-trained solely on the synthetic dataset corresponding to the target real dataset. Subsequently, the model is fine-tuned using the same approach as Strategy B. This strategy assesses the impact of training DANIEL on multiple synthetic datasets concurrently.

For each strategy, training continues with a partial reliance on synthetic data while real data is incrementally introduced. The probability of incorporating real data begins at 0\% and increases to 80\% throughout 300,000 steps.

\subsection{Evaluation metrics}
To assess our model's performance in HTR, we compute both the Character Error Rate (CER) and the Word Error Rate (WER) at the page level. When computing CER and WER, special tokens such as layout and named entity tokens are excluded.

Regarding the layout recognition evaluation, we use the Layout Ordering Error Rate (LOER) and the mean Average Precision on CER ($\mathrm{mAP_{CER}}$) introduced in \cite{coquenet_dan}.

Additionally, for the evaluation of NER performance, we utilize the F1 score, computed using the NERval library \cite{nerval2021}.
This approach matches predictions to their ground truths at the character level with an acceptable CER threshold of 30\% to determine a valid match between the predicted and actual data.

\subsection{Results}
\subsubsection{Handwritten Text Recognition performance}\label{htr-results}

For each dataset, we present the results obtained for each fine-tuning strategy described in subsection \ref{fine-tuning-strategies}.
The evaluation results for the IAM dataset are detailed in Table \ref{tab:tr-iam}.

DANIEL achieves very competitive results, outperforming Dessurt in terms of CER with a CER of 4.38\% and surpassing the DAN in terms of WER with a WER of 10.89\% using fine-tuning strategy B.
As observed in \cite{davis_end--end_2022}, the gap between the CER and WER is smaller for subword-based methods than for character-based ones.
This characteristic highlights the direct contribution and advantages of large subword language models.

Pre-training on multiple synthetic datasets proves beneficial for the IAM dataset, even if the layouts of these synthetic datasets differ. Strategies A, B, and C outperform strategy D, likely because the inclusion of text from different languages in the synthetic datasets acts as regularization, reducing the risk of overfitting on the synthetic data.
From the results of strategies A, B, and C, we can conclude
that the best performance is achieved when no other datasets except IAM are used during training. This phenomenon might be explained by the relatively large size and simple layout (paragraphs) of the IAM dataset. Thus, training exclusively on IAM is sufficient in terms of diversity and does not lead to overfitting.
Consequently, the model does not benefit from including documents from other real datasets,
while using other real data increases the risk of the model not being sufficiently specialized. 
If the real data used for training do not closely align with the evaluation data, this mismatch can adversely affect performance.

\begin{table}[h]
\centering
\caption{Text Recognition (TR) results on the IAM dataset (RWTH split). Metrics are expressed in percentages.}
\label{tab:tr-iam}
\begin{tabular}{c|cc}
\hline
Method & CER & WER \\ \hline
OrigamiNet \cite{yousef_origaminet} & 4.7 & -  \\ 

VAN \cite{coquenet_end} & 4.5 & 14.6\\ 
DAN \cite{tarride_key-value_2023} & \textbf{4.3} & 13.66 \\ 
Dessurt \cite{davis_end--end_2022} & 4.8 & \textbf{10.2} \\ 
\hline

Ours - strategy A & 5.02 & 11.95 \\
Ours - strategy B & 4.38 & 10.89 \\
Ours - strategy C & 4.75     & 11.44 \\
Ours - strategy D & 5.98 & 13.98     \\ \hline
\end{tabular}%
\end{table}

Table \ref{tab:tr-rimes} presents the evaluation of DANIEL on the RIMES 2009 dataset.
This time, fine-tuning strategy C achieves state-of-the-art results
in terms of WER with a WER of 11.22\% and a CER of 5.8\%.

The performance is lower compared to the DAN
in terms of $\mathrm{mAP_{CER}}$ and LOER.
These results indicate that some errors contributing to DANIEL's CER and WER are due to reading order issues, such as the inversion of the order between two paragraphs in the prediction. 

This discrepancy arises from the synthetic data used, which includes text from Wikipedia rather than mail text, as found in the real data. 
Consequently, the model can only rely on the spatial positions of text blocks and not on the semantic content.
Using a text corpus with a semantic field closer to the real data could therefore improve performance in layout recognition.

Similar to the IAM dataset, multilingual strategies A, B, and C outperform the monolingual strategy D.
However, the gap in CER between multilingual strategies is smaller. This is likely because the RIMES 2009 layout is more complex than that of IAM, making training on other complex datasets, such as READ 2016, more beneficial for RIMES 2009 than for IAM.

\begin{table}[h]
\centering
\caption{Text Recognition (TR) results on the RIMES 2009 dataset. Metrics are expressed in percentages.
}
\label{tab:tr-rimes}
\begin{tabular}{ccccc}
\hline
Method & CER & WER & LOER & $\mathrm{mAP_{CER}}$ \\ \hline
DAN \cite{coquenet_dan} & \textbf{4.54} & 11.85 & \textbf{3.82} & \textbf{93.74}\\ 
Faster-DAN \cite{coquenet_faster_2023} & 6.38 & 13.69 & 4.48 & 91.00\\ \hline
Ours - strategy A & 6.01 & 11.40 & 5.50 & 90.13\\
Ours - strategy B & 5.81 & 11.24 & 4.60 & 91.99\\
Ours - strategy C & 5.80 &  \textbf{11.22} & 6.57 & 90.21\\ 
Ours - strategy D & 6.43 & 12.04 &5.26 & 89.44\\ \hline
\end{tabular}%
\end{table}

The results for READ 2016 are presented in Table \ref{tab:tr-read}. A clear distinction can be observed between fine-tuning methods based on multilingual pretraining and strategy D, which includes only synthetic and real data from READ 2016.
Using strategy C, DANIEL achieves competitive results with a CER of 4.03\% and a WER of 15.63\%.
Given the limited amount of training data in READ 2016, the inclusion of synthetic and real data from other datasets appears to assist the model by providing regularization,
even if the additional data are not highly similar to those of READ 2016.
Further improvements could be achieved by using synthetic data more visually similar to READ 2016, such as the synthetic data introduced in \cite{barrere_training_2024}.

\begin{table}[h]
    \centering
    \caption{Text Recognition (TR) results on the READ 2016 dataset. Metrics are expressed in percentages.}
    \label{tab:tr-read}
        \begin{tabular}{ccccc}
            \hline
            Method & CER & WER & LOER & $\mathrm{mAP_{CER}}$ \\ \hline
            DAN \cite{coquenet_dan}& \textbf{3.43} & \textbf{13.05} & 5.17 & 93.32\\ 
            Faster-DAN \cite{coquenet_faster_2023} & 3.95 & 14.06 & 3.82 & \textbf{94.20}\\ \hline
            Ours - strategy A & 4.41 & 16.03 & 4.95 & 91.42 \\
            Ours - strategy B & 4.34 & 15.54 & 3.49 & 90.80\\
            Ours - strategy C & 4.03 & 15.63 & \textbf{3.37} & 92.66 \\
            Ours - strategy D & 5.41 & 19.27 & 4.95 & 87.47\\ \hline
        \end{tabular}%
\end{table}

The final HTR dataset on which DANIEL was evaluated is M-POPP, with detailed results presented in Table \ref{tab:tr-popp}. Using fine-tuning strategy C, DANIEL achieved state-of-the-art performance for HTR with a CER of 5.72\% and a WER of 14.08\%. Additionally, it set new benchmarks for layout recognition with a LOER of 1.34\% and a $\mathrm{mAP_{CER}}$ of 89.28\%. These results demonstrate that our training scheme enables DANIEL to achieve excellent performance even on a small and challenging dataset like M-POPP.

Regarding fine-tuning strategies, the variability in results across different approaches is more pronounced for M-POPP compared to other datasets.
The disparity in $\mathrm{mAP_{CER}}$ between strategy A and strategies B and C indicates that the increased CER for strategy A is likely due to errors in layout analysis. An analysis of the predictions reveals that DANIEL occasionally misses elements in the margin, significantly increasing the CER in the affected images. This may be because the model is disrupted by the diverse layouts of other real datasets it has encountered during training.
Training DANIEL on M-POPP using the monolingual strategy D resulted in highly degraded performance, underscoring the importance of multilingual pre-training.

Given the recent introduction of this dataset, there is likely still room for improvement. For instance, the text corpus used to generate the synthetic data may be sub-optimal. Wikipedia provides a very generalist language model, whereas M-POPP corresponds to a very specific linguistic context. Therefore, using a text corpus more similar to the marriage records in M-POPP could improve the results.
Another potential solution could be to apply a self-training process, using predictions on the unlabelled images from the M-POPP corpus as pseudo-labels. This method has
been successfully applied to handwritten census tables in \cite{constum_recognition_2022}.

\begin{table}[h]
\centering
\caption{Text Recognition (TR) results on the M-POPP dataset. Metrics are expressed in percentages. TR+IE stands for joint Text Recognition and Information Extraction.}
\label{tab:tr-popp}
\resizebox{\columnwidth}{!}{%
\begin{tabular}{ccccc}
\hline
Method & CER & WER & LOER & $\mathrm{mAP_{CER}}$ \\ \hline
DAN - TR \cite{constum2024endtoend} & 7.21 & 16.42 & 5.35 & 83.03 \\ 
DAN - TR+IE \cite{constum2024endtoend} & 6.52 & 14.80 & 3.79 & 86.29\\ \hline
Ours - strategy A & 8.35 & 17.36 & 5.12 & 80.98\\
Ours - strategy B & 6.61 & 15.33 & 6.01 & 84.41 \\
Ours - strategy C & \textbf{5.72} &  \textbf{14.08} & \textbf{1.34} & \textbf{89.28}\\ 
Ours - strategy D & 14.28 & 25.47 & 3.56 & 67.10\\ \hline
\end{tabular}%
}
\end{table}

\textbf{Summary - HTR}
Based on the results obtained from the different datasets, it appears that the best fine-tuning strategies for HTR are strategies B and C. Strategy C seems preferable in the case of datasets like READ 2016 or M-POPP, which contains little training data and are relatively different from the synthetic data. For the other datasets, it is preferable to use strategy B since this strategy involves fewer steps.

\subsubsection{Named Entity Recognition performance}

Besides HTR, we also evaluate the DANIEL capabilities on NER tasks.
The results are detailed in tables \ref{tab:ie-iam} and \ref{tab:ie-popp} for IAM NER and M-POPP NER respectively.

On the IAM NER dataset, DANIEL achieves 
very impressive performances with new state-of-the-art results.
On the RWTH split DANIEL obtains an F1 score of 50.06\% with fine-tuning strategy C. The only method that performs better is obtained by \cite{tuselmann_are_2021}. 
It should be noted that this method is a sequential approach based first on a text recognition model and then on a BERT language model specially trained on the annotation of named entities from the OntoNotes v5 ontology. Moreover, their text recognition model works on word images which avoids any reading order problems. If only end-to-end methods are considered, DANIEL is thus state-of-the-art on the RWTH split.
On the custom split, the results are even more impressive, with DANIEL achieving a new state-of-the-art F1 score of 58.29\%, outperforming even sequential methods. This performance difference between the two splits can be attributed to their 
design.
DANIEL performs better in NER on the custom split because
this split was specifically designed for the NER task when this task was introduced on IAM
whereas the RWTH split was designed solely for handwriting recognition purposes.
However, as Tusselman's method uses a BERT model pre-trained for NER, this method shows less dependency on the distribution of named entities in the real data.
This explains why Tusselman's results are more consistent across both splits.
Regarding the fine-tuning strategies, we observe that the multilingual strategies (A, B, and C) outperform the monolingual strategy D.
However, the performance differences among the multilingual strategies are relatively small.
This is likely because the other datasets include only text recognition tasks and lack named entity annotations, which is therefore less useful for IAM NER.

Regarding M-POPP NER, the DANIEL model achieves a state-of-the-art F1 score of 76.37\% using fine-tuning strategy C. 
The comparison between the monolingual and multilingual strategies indicates that training on data containing named entities enhances NER performance on the target dataset, even when these named entities do not conform to the same ontology. However, we believe that DANIEL's performance on M-POPP NER could be further improved by incorporating synthetic data that includes named entities from the M-POPP dataset. 
Again, self-training on M-POPP NER might also be a viable approach to explore,
as it would provide more training data adhering to the same named entity ontology.

\textbf{Summary - NER}
Based on the results from both the IAM NER and M-POPP NER datasets, fine-tuning strategies B and C demonstrate the best performance for NER. Similar to handwriting recognition, the choice between these two strategies depends on the characteristics of the target dataset. For small datasets lacking synthetic data with named entities, strategy C is preferable as it leverages real data from other datasets with named entities. For other datasets, both strategies B and C yield comparable results; however, strategy B is preferable due to its simpler implementation and the absence of a requirement for real data from other datasets.

\begin{table}[h]
\centering
\caption{Information Extraction (IE) results on the IAM NER dataset for both RWTH and custom splits. Metrics are expressed in percentages.}
\label{tab:ie-iam}
\begin{tabular}{cccc}
\hline
Method & \begin{tabular}[c]{@{}c@{}}F1\\ (RWTH)\end{tabular} & \begin{tabular}[c]{@{}c@{}}F1\\ (custom)\end{tabular} & Type\\ \hline
Toledo \cite{toledo} & 14.9 & 18.0 & Sequential\\
Rowtula \cite{rowtula} & 32.3 & 30.3 & Sequential\\
Tusselman \cite{tuselmann_are_2021} & \textbf{52.0} & 53.6 & Sequential\\ \hline
Dessurt \cite{davis_end--end_2022} & 40.4 & 48.5 & End-to-end\\
DAN \cite{tarride_key-value_2023}& 31.3 & - & End-to-end\\ \hline
Ours - strategy A & 46.35 & 54.83 & End-to-end\\
Ours - strategy B & 48.86 & \textbf{58.29} & End-to-end\\
Ours - strategy C & 50.06 & 57.55 & End-to-end\\
Ours - strategy D & 41.79 & 53.25 & End-to-end\\ \hline
\end{tabular}%
\end{table}

\begin{table}[h]
\centering
\caption{Information Extraction (IE) results on the M-POPP NER dataset. Metrics are expressed in percentages.}
\label{tab:ie-popp}
\begin{tabular}{ccc}
\hline
 &  F1 \\ \hline
DAN \cite{constum2024endtoend} & \textbf{76.37} \\ \hline
Ours - strategy A   &  71.71 \\
Ours - strategy B   &  74.20\\
Ours - strategy C   &  \textbf{76.37}\\
Ours - strategy D   & 59.64 \\ \hline
\end{tabular}%
\end{table}

\subsubsection{Inference speed}
If DANIEL shows very interesting performances in text recognition and named entity extraction, it also excels in terms of inference speed. This parameter, regularly mentioned in VDU literature, is often overlooked in text recognition and named entity extraction articles applied to handwritten documents. However, inference speed is a crucial factor for real-world applications, especially when it comes to processing documents in real-time or in very large quantities.
Therefore, in this section, we compare the inference time of different state-of-the-art 
HTR and HTR+IE systems.
For practical reasons, we have only selected methods whose code and weights were publicly available. Moreover, we have also discarded methods applied to images of lines or paragraphs because their inference time does not take into account the time needed for the segmentation steps they require. 
For HTR, the selected methods are DAN, Faster-DAN, Dessurt and DANIEL.
For NER, only the methods based on DAN meet the required criteria. Indeed, while pre-trained Dessurt weights are publicly available for handwriting recognition,
they are not available for NER.

Before assessing the inference speed of the selected methods, it is relevant to first examine the size of each model. Table \ref{tab:nb-parameters} presents the number of parameters for each model. 
Models operating at the subword level have significantly more parameters compared to those working at the character level. This discrepancy is primarily due to the size of the vocabulary, which is considerably larger for subword-level recognition models.
As detailed in \ref{vocab}, the vocabulary size constitutes a substantial portion of the total number of parameters in subword-based models.
Nevertheless, a higher parameter count does not inherently correlate with a reduction in inference speed, as demonstrated in the next part of this study.

\begin{table}[h]
\centering
\caption{Number of parameters for DAN, Faster-DAN, Dessurt and DANIEL. Parameters are expressed in millions.}
\label{tab:nb-parameters}
\begin{tabular}{ccccc}
\hline
Model & DAN & Faster-DAN & Dessurt & DANIEL\\ \hline
\# params & 7.0M & 7.0M & 163.9M & 154.0M\\ \hline
\end{tabular}%
\end{table}

Table \ref{tab:inference-speed} shows a comparison of the prediction times between DAN, Faster-DAN, Dessurt and DANIEL for HTR.
These prediction times correspond to the average time per image computed over the prediction on the complete test set of each dataset with a batch size of 1 on an A100 GPU. 
We can observe that one of the greatest assets of DANIEL is its inference speed.
Indeed, DANIEL is faster than every other existing model while showing competitive or better recognition performance.
For instance, DANIEL is at least 4.8 faster than DAN.
The inference speed difference between DANIEL and DAN is likely due to DANIEL's prediction process, which occurs at the subword level rather than the character level. Thus, DANIEL predicts fewer tokens than DAN to recognize the same text from a given document.
Additionally, the superior speed of DANIEL comes from its implementation relying on the Hugging Face transformers library\footnote{\url{https://github.com/huggingface/transformers}}, which is renowned for its efficiency.
Although DANIEL and Dessurt are both based on subword prediction and have a similar number of parameters,
DANIEL is four times faster than Dessurt on IAM. 
We hypothesize that this difference is due to the architectural differences between the two models. Indeed, Dessurt is entirely based on transformer layers, while DANIEL uses a fully convolutional encoder.
\begin{table}[h]
\centering
\caption{Inference speed results for HTR on RIMES 2009, READ 2016, IAM and M-POPP. Results are expressed in seconds per image.}
\label{tab:inference-speed}
\resizebox{\columnwidth}{!}{%
\begin{tabular}{ccccc}
\hline
Method & \begin{tabular}[c]{@{}c@{}}RIMES\\ 2009\end{tabular} & \begin{tabular}[c]{@{}c@{}}READ\\ 2016\end{tabular} & IAM & M-POPP\\ \hline
DAN \cite{coquenet_dan} & 4.94 & 3.84 &  2.85 & 15.5\\ 
Faster-DAN \cite{coquenet_faster_2023} & 0.99 & 0.84 & - & -\\ 
Dessurt \cite{davis_end--end_2022} & - & - & 1.48 & -\\ \hline
Ours & \textbf{0.77} & \textbf{0.80} & \textbf{0.37} & \textbf{2.66} \\ \hline
\end{tabular}%
}
\end{table}

Table \ref{tab:inference-speed-ner} presents a comparative analysis of prediction times between DAN and DANIEL for HTR combined with NER, using the same protocol as employed for HTR.
Once again, DANIEL demonstrates faster inference speed for the HTR+NER task compared to DAN.
The DANIEL is indeed 6.35 times faster than the DAN on IAM and 5.68 times faster on M-POPP.

\begin{table}[h]
\centering
\caption{Inference speed results for NER on IAM NER and M-POPP NER. Results are expressed in seconds per image.}
\label{tab:inference-speed-ner}
\begin{tabular}{ccccc}
\hline
Method & IAM NER & M-POPP NER\\ \hline
DAN \cite{coquenet_dan} & 3.24 & 17.94\\ 
Ours & \textbf{0.51} & \textbf{3.16}\\ \hline
\end{tabular}%
\end{table}

\section{Further studies}
\subsection{Encoder comparison}
\label{encoder-comparison}
In this section, we study different convolutional encoder architectures. 
First, we aim
to assess the relevance of choosing an encoder architecture derived from DAN. In this respect, we compare the performance of the chosen encoder with a ConvNext v2 encoder \cite{woo2023convnext}.
Indeed, this kind of architecture currently achieves the best performance on ImageNet among convolutional architectures. To ensure a fair comparison, we designed a ConvNext v2 architecture with a similar number of parameters as the DANIEL encoder, which is 20.0M.
Therefore, we build on the ConvNext v2 \textit{femto} encoder
and we set the output embedding size of the last block to 1024 so that this encoder is compatible with the input embedding size of the DANIEL decoder.
Additionally, we modify the strides of two convolutional layers so that the output feature map size is \( H_f = \frac{H}{32} \), \( W_f = \frac{W}{8} \) following the notation of \ref{presentation-encoder}.
Thus, the resulting architecture has 19.8M parameters.

Secondly, we show that the
scale of the chosen encoder is relevant when combined with the DANIEL decoder.
Therefore, we compare the DANIEL encoder with two other variants. The first model named S is a small variant with 4.0M parameters, and the second model named L in a larger variant with 44.2M parameters. Table \ref{tab:encoder-comparison} gives a detailled comparison of DANIEL encoder with these two variants. 

\begin{table}[]
\centering
\caption{Details of the output embedding size for each block of the evaluated variants of the DAN encoder.}
\label{tab:encoder-comparison}
\begin{tabular}{cccc}
\hline
\begin{tabular}[c]{@{}c@{}}Block\\ number\end{tabular} & Encoder S & \begin{tabular}[c]{@{}c@{}}DANIEL\\ encoder\end{tabular} & Encoder L \\ \hline
1 & 16 & 32 & 32 \\
2 & 32 & 64 & 64 \\
3 & 64 & 128 & 128 \\
4 & 128 & 256 & 256 \\
5 & 128 & 512 & 512 \\
6 & 128 & 512 & 1024 \\
7 & 128 & 512 & 1024 \\
8 & 128 & 512 & 1024 \\
9 & 256 & 512 & 1024 \\
10 & 1024 & 1024 & 1024 \\ \hline
\# params & 4.0M & 20.0M & 44.2M \\ \hline
\end{tabular}%
\end{table}

To limit the training time, we train each model according to the monolingual D training strategy on the IAM dataset. The performance of each encoder is detailed in Table \ref{tab:encoder-results}.
Based on the results, we can conclude that using an encoder architecture derived from the DAN encoder is indeed effective. The DANIEL encoder achieves superior results with a CER of 5.98\% and a WER of 13.98\%, compared to a CER of 6.36\% and a WER of 14.26\% for the ConvNext v2 encoder.
Since the performances of these two encoders are quite similar, further investigation into using an encoder from the ConvNext model family could be valuable in future research.

In terms of encoder size, the results indicate that an encoder with too few parameters is unsuitable for a large decoder like the one of DANIEL, as evidenced by the S encoder's CER of 10.85\% and WER of 22.28\%. Additionally, using an encoder larger than that of DANIEL is not advisable, as it does not enhance performance and only increases the number of parameters.

\begin{table}[h]
\centering
\caption{Text Recognition (TR) results on the IAM dataset (RWTH split) for each of the evaluated encoders. Metrics are expressed in percentages.}
\label{tab:encoder-results}
\begin{tabular}{c|cc}
\hline
Encoder & CER & WER \\ \hline
Encoder S & 10.85 & 22.28 \\
DANIEL encoder & \textbf{5.98} & \textbf{13.98}\\
Encoder L & 6.50 & 14.72\\
ConvNext v2 encoder & 6.36 & 14.26\\ \hline
\end{tabular}%
\end{table}

\section{Conclusion}
This research introduced the Document Attention Network for Information
Extraction and Labelling (DANIEL), an end-to-end architecture that has shown remarkable performance in the domain of handwritten document understanding. Our approach, which blends a convolutional encoder with an autoregressive transformer decoder, has demonstrated its efficacy across multiple datasets on Handwritten Text Recognition and Named Entity Recognition, setting new benchmarks for these tasks.

Unlike current VDU architectures that require a fixed input size due to their vision-transformer encoder, DANIEL, with its convolutional encoder, can accommodate input images of any size or aspect ratio without resizing. This is a critical advantage for handwritten document recognition, given the variability in document and character sizes.

DANIEL achieves
a new state-of-the-art performance
on RIMES 2009 and M-POPP and competitive results on IAM and READ 2016 for Handwriting Text Recognition.
For Named Entity Recognition, DANIEL achieves a new state-of-the-art performance on IAM NER and state-of-the-art results on M-POPP NER.
We demonstrate that an end-to-end architecture such as DANIEL can surpass every sequential method for NER on IAM NER including methods using language models such as BERT. 
This is made possible through an innovative model distillation method that allows for the transfer of knowledge from a language model.

Not only has DANIEL proven to be competitive in performance metrics, but it has also achieved remarkable inference speed, surpassing existing architectures in inference efficiency. The synthesis of subword-scale prediction and optimized implementation positions DANIEL as a leading model for real-world applications where speed and accuracy are paramount.

Moreover, the introduction of a novel
synthetic data generator specialized for handwritten documents and tailored for diverse layouts and languages has enhanced DANIEL's capacity to grasp and process multiple datasets.
Our findings indicate that pre-training DANIEL on synthetic data across various languages and layouts improves its performance in both HTR and NER. Additionally, incorporating real data from other datasets is recommended when dealing with small datasets.

While DANIEL marks a significant advancement, we also recognize the scope for further refinement. 
Future work may delve into enhancing its generalization capabilities for new layouts or new information categories to extract without designing new synthetic data.
An intriguing avenue for future research could be exploring self-training methods. This approach involves using model predictions on unlabeled data as pseudo-labels, thereby increasing the volume of training data.

\section{Acknowledgements}
This work was performed using HPC resources from GENCI-IDRIS (Grant 2023-AD011013149R2). This work was supported by the Normandy regional council and the ANR EXO-POPP project, grant ANR-21-CE38-0004 (French Agence Nationale de la Recherche).

\appendix

\section{Details on synthetic data}
We have observed significant variations in the size of characters generated by different fonts, even when the same font size is specified. Given that our synthetic data generator already varies the font sizes in the images it creates, it is crucial to maintain consistency across images with identical font sizes. 
To achieve this, we standardize the fonts based on size. Specifically, we generate a consistent test phrase for each font, and from this, we calculate a normalization factor for each font to ensure the resulting images have comparable widths.

\section{Synthetic data examples}

\begin{figure}[h]
    \centering
    \fbox{\includegraphics[width=6cm]{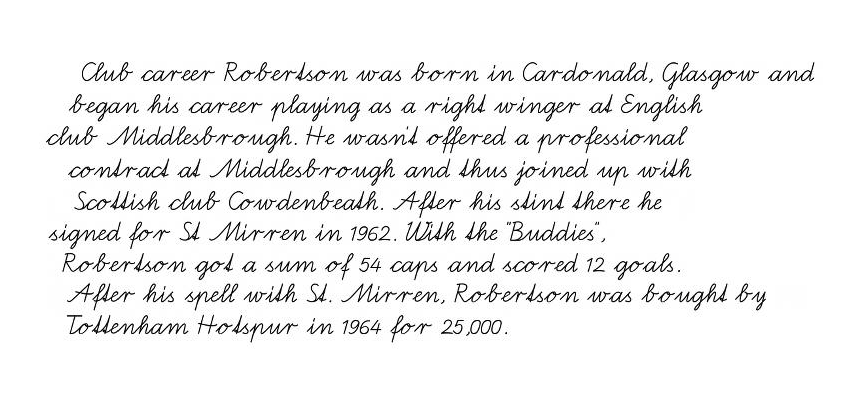}}
    \caption{Example of synthetic data for IAM.}
    \label{fig:exemple-iam-synth}
\end{figure}

\begin{figure}[h]
    \centering
    \fbox{\includegraphics[width=0.22\textwidth]{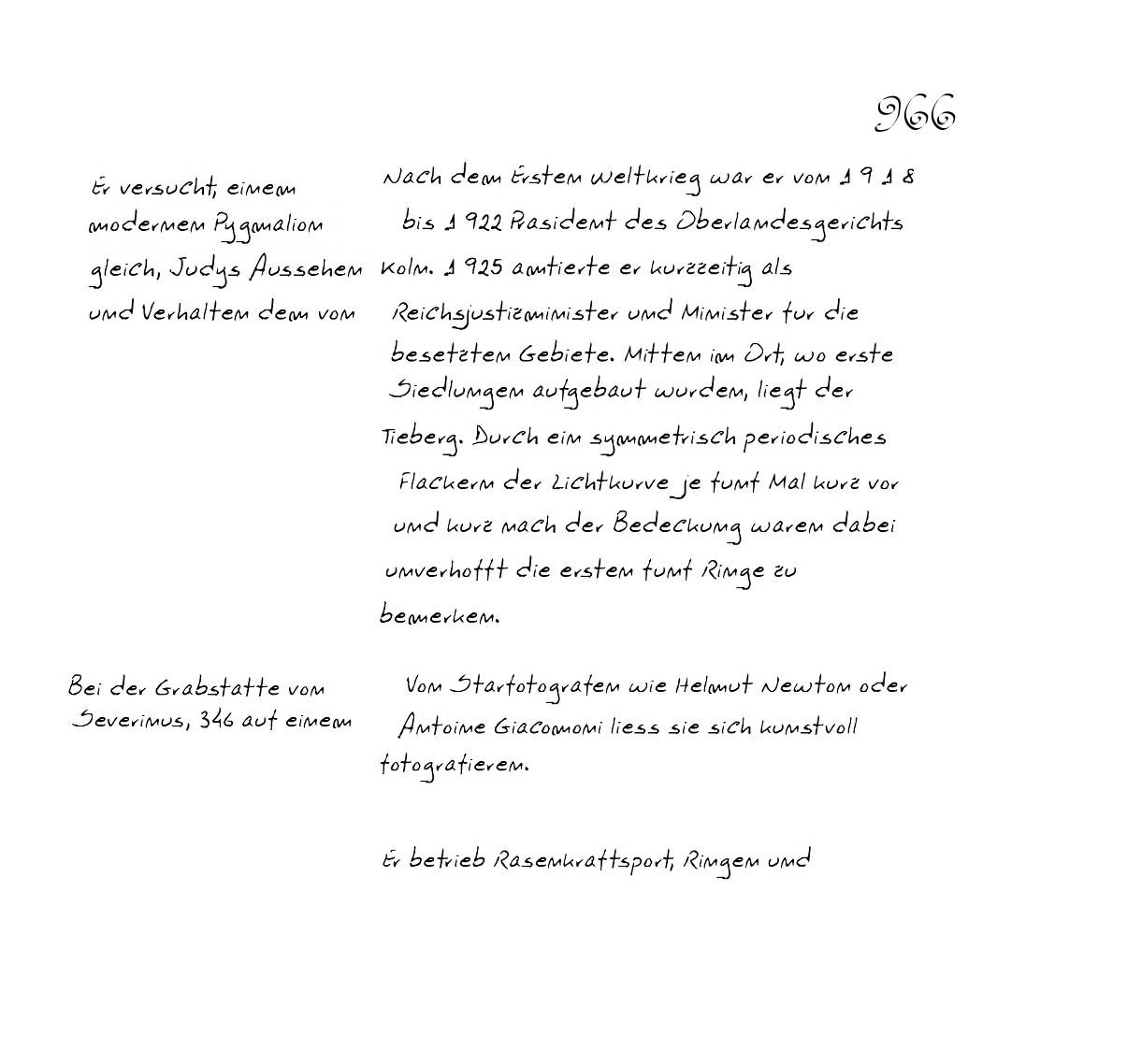}}
    \fbox{\includegraphics[width=0.22\textwidth]{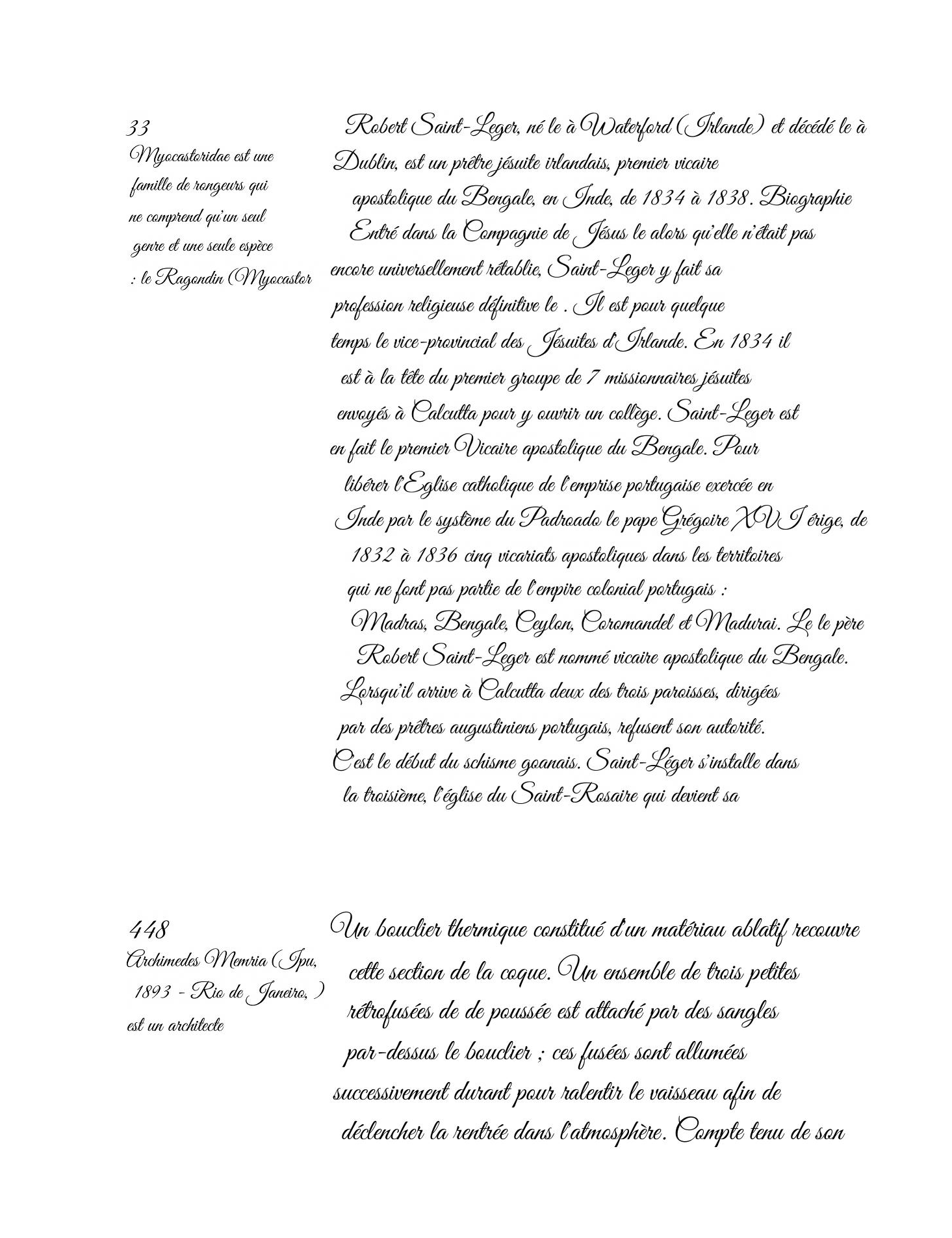}}
    \caption{Examples of synthetic data for READ 2016 (left) and M-POPP (right) datasets.}
    \label{fig:both-images-synth-read-exopopp}
\end{figure}

We provide examples of synthetic data for the dataset IAM in Figure \ref{fig:exemple-iam-synth}, and datasets READ 2016 and M-POPP in Figure \ref{fig:both-images-synth-read-exopopp}.

\section{Details of implementation}\label{vocab}
\subsection{Vocabulary Reduction}
The tokenizer of DANIEL 
is based on the tokenizer from Donut whose vocabulary is a pruned version of mBART \cite{liu_multilingual_2020} vocabulary\footnote{As explained in \url{https://github.com/hyunwoongko/asian-bart}}. 
We envisioned DANIEL as a generalist model capable of adapting to all documents in languages that use the Latin alphabet. This means that some of the subwords present in the Donut's tokenizer vocabulary will never be encountered by DANIEL. Indeed, it contains subwords including the Latin alphabet but also sinograms, the Arabic alphabet, Cyrillic, etc. However, the size of the vocabulary directly influences two layers in the decoder:
\begin{itemize}
    \item The decision layer, which allows transitioning from an embedding of size 1024 to output probabilities over the 
    57718
    subwords of the vocabulary.
    \item The embedding layer, which enables encoding each of the
    57718
    subwords into an embedding vector of size 1024.
\end{itemize}

These two layers are dense layers and each comprises approximately
$1024\times57718 \approx59.1$
M parameters, totaling
118.2M
parameters. If we count the weights of these layers in the total number of parameters of the DANIEL without refinement of vocabulary, it totals
206.3M
parameters. This means that these two layers represent
57.3\%
of the parameters of this model.

For this reason, we decided to reduce the size of DANIEL vocabulary.
To decide which subwords we wanted to remove, we relied on the Unicode character table\footnote{\url{https://en.wikipedia.org/wiki/List\_of\_Unicode\_characters}}, where characters are grouped by blocks, and these blocks often correspond to the original language.
We thus established the list of blocks (and therefore characters) that we did not want to support. Then we considered that any subword containing at least one of the unsupported characters was rejected.

To ensure that we were not rejecting a subword that might be used, we compiled a list of subwords encountered in all the real and synthetic data used.
In total in the vocabulary of the Donut's tokenizer:
27610
subwords are used and 25565 subwords have been rejected. 
The intersection between the rejected subwords and the used subwords is empty. 
Finally,
4543
subwords are neither rejected nor used.
Some might be encountered later and others will never be used but belong to Unicode blocks where the superfluous aspect is not obviously apparent.
The vocabulary size goes from
57718
to
32153,
reducing its size by
44\%.
The embedding and decision layers thus see their size reduced from
59.1M
to
$1024\times32153\approx32.9M$
parameters each.
The number of DANIEL parameters thus drops from 206.3M to 154.0M.

\bibliographystyle{unsrt}

\end{document}